\documentclass[lettersize,journal]{IEEEtran}
\usepackage{algorithm}
\usepackage{caption}
\usepackage{makecell}
\usepackage{subcaption}
\usepackage{array}
\usepackage{textcomp}
\usepackage{stfloats}
\usepackage{url}
\usepackage{verbatim}
\usepackage{graphicx}
\usepackage{cite}
\usepackage{amsmath,amssymb,amsfonts}
\usepackage{algorithmic}
\usepackage{times}
\usepackage{latexsym}
\usepackage{makecell}

\usepackage{here}
\usepackage{stmaryrd}
\usepackage{multirow}
\usepackage{changepage}
\usepackage{threeparttable}
\usepackage{tabu}
\usepackage{longtable}
\usepackage{float}
\usepackage{colortbl}
\usepackage[dvipsnames,table]{xcolor}
\usepackage{tabularx}
\usepackage{array}
\usepackage[utf8]{inputenc} 
\usepackage[T1]{fontenc}    
\usepackage[hidelinks]{hyperref}       
\usepackage{booktabs}       
\usepackage{nicefrac}       
\usepackage{microtype}      
\usepackage{lipsum}
\usepackage{fancyhdr}

\begin{document}

\title{Multimodal Detection of Bots on X (Twitter) using Transformers}

\author{Loukas Ilias, Ioannis Michail Kazelidis, Dimitris Askounis
\thanks{The authors are with the Decision Support Systems Laboratory, School of Electrical and Computer Engineering, National Technical University of Athens, 15780 Athens, Greece (e-mail: lilias@epu.ntua.gr; gkazelid@gmail.com; askous@epu.ntua.gr).}
}

\markboth{Journal of \LaTeX\ Class Files,~Vol.~14, No.~8, August~2021}%
{Shell \MakeLowercase{\textit{et al.}}: A Sample Article Using IEEEtran.cls for IEEE Journals}


\fancypagestyle{plain}{
  \renewcommand{\headrulewidth}{0pt}%
  \fancyhf{}
  \fancyfoot[C]{\thepage}%
}
\fancypagestyle{firstpage}{
  \renewcommand{\headrulewidth}{0pt}%
  \renewcommand{\footrulewidth}{-0.2pt}%
  \fancyhf{}
  \fancyfoot[C]{\tiny © 2024 IEEE. Personal use of this material is permitted. Permission from IEEE must be obtained for all other uses, in any current or future media, including reprinting/republishing this material for advertising or promotional purposes, creating new collective works, for resale or redistribution to servers or lists, or reuse of any copyrighted component of this work in other works.}%
}

\maketitle

\begin{abstract}
Although not all bots are malicious, the vast majority of them are responsible for spreading misinformation and manipulating the public opinion about several issues, i.e., elections and many more. Therefore, the early detection of bots is crucial. Although there have been proposed methods for detecting bots in social media, there are still substantial limitations. For instance, existing research initiatives still extract a large number of features and train traditional machine learning algorithms or use GloVe embeddings and train LSTMs. However, feature extraction is a tedious procedure demanding domain expertise. Also, language models based on transformers have been proved to be better than LSTMs. Other approaches create large graphs and train graph neural networks requiring in this way many hours for training and access to computational resources. To tackle these limitations, this is the first study employing only the user description field and images of three channels denoting the type and content of tweets posted by the users. Firstly, we create digital DNA sequences, transform them to 3d images, and apply pretrained models of the vision domain, including EfficientNet, AlexNet, VGG16, etc. Next, we propose a multimodal approach, where we use TwHIN-BERT for getting the textual representation of the user description field and employ VGG16 for acquiring the visual representation for the image modality. We propose three different fusion methods, namely concatenation, gated multimodal unit, and crossmodal attention, for fusing the different modalities and compare their performances. Finally, we present a qualitative analysis of the behavior of our best performing model. Extensive experiments conducted on the Cresci'17 and TwiBot-20 datasets demonstrate valuable advantages of our introduced approaches over state-of-the-art ones.  
\end{abstract}


\section{Introduction}
\thispagestyle{firstpage}

Social media platforms, such as Twitter (rebranded to X in 2023) and Reddit, constitute a valuable form of information, where the users have the opportunity to express their feelings, communicate with people around the world, and get informed about the news. For instance, social media have been used for the early detection of mental disorders \cite{10154134}. Especially in Twitter, people can be informed about everything happening at the moment through the Twitter trends. However, social media are often manipulated by bots. Although not all bots are malicious and dangerous, the majority of them constitute the main form of misinformation \cite{cresci2020detecting}. Research has showed that bots can influence elections \cite{bessi2016social, brachten2017strategies}, promote phishing attacks \cite{7841038}, and constitute an effective tool for manipulating social media by spreading articles of low-credibility scores \cite{shao2018spread}. According to \cite{doi:10.1177/20563051221104749}, bots dominated hate speech during COVID-19. Therefore, the early detection of bots appears to be imperative nowadays. 

Existing research initiatives train shallow machine learning classifiers for recognizing bots in Twitter and propose feature extraction approaches \cite{ILIAS2021107360,ouni2022bots}. However, feature extraction constitutes a time-consuming process, demands some level of domain expertise, while the optimal feature set for each specific task may not be found. In addition, there have been introduced deep learning approaches. However, these approaches either use as input a large number of features, including user and tweet metadata, or exploit GloVe embeddings and train LSTM neural networks \cite{KUDUGUNTA2018312,hayawi2022deeprobot}. However, fine-tuning language models based on transformers requires less time for training and yields better evaluation results. Additionally, methods fine-tuning BERT, RoBERTa, etc. models cannot be applied to other languages except the english one. Moreover, graph-based approaches have been introduced for detecting bots. However, feature extraction techniques are also applied for creating feature vectors per users, which represent the nodes in graphs \cite{10.1145/3487351.3488336}. At the same time, large graphs are often created, which require a lot of time for training and require access to computational resources.

To address these limitations, we present the first study which exploits only the user description and the sequence of actions that an account performs for recognizing  bots in Twitter. In the current study, we introduce both unimodal, i.e., neural networks exploiting either text or image, and multimodal approaches, i.e., neural networks utilizing both text and image. First, motivated by \cite{7436643,7876716}, we design the digital DNA per user, which indicates the sequence of actions that an account performs. Next, we adopt the methodology proposed by \cite{10.1007/978-3-031-35995-8_42} for converting the DNA sequence into an image. After that, we fine-tune several pretrained models, including AlexNet, ResNet, VGG16, etc., and compare their performances. For recognizing bots through user descriptions, this is the first study fine-tuning TwHIN-BERT \cite{zhang2022twhin} in the task of bot detection in Twitter. Regarding the multimodal approaches, we employ VGG16, which constitutes our best performing model, and extract the visual representation. In terms of the textual modality, we employ the TwHIN-BERT and extract the textual representation. Then, we propose three methods for fusing the representations of the different modalities. Specifically, first we concatenate the representation vectors of the two modalities. Second, we exploit a gated multimodal unit (GMU), which controls the importance of each modality towards the final classification. Thirdly, we use a cross-attention mechanism for capturing the inter-modal interactions. Experiments conducted on the Cresci'17 and TwiBot-20 datasets show that the introduced cross-modal model outperforms the competitive multimodal ones.

Our main contributions can be summarized as follows:

\begin{itemize}
    \item We create digital DNA sequences per user based on both the type and contents of the tweets and transform these sequences into 3d images.
    \item We employ and compare many models of the vision domain, including EfficientNet, AlexNet, VGG16, etc. by utilizing the DNA sequence as an image consisting of three channels.
    \item This is the first study fine-tuning the TwHIN-BERT model for recognizing bots utilizing only the user description field.
    \item To the best of our knowledge, this is the first study introducing multimodal models employing the user description and the representation of the DNA sequence as an image.
    \item We present the first study utilizing a gated multimodal unit and cross-attention mechanism, and comparing the fusion methods.
\end{itemize}

\section{Related Work}

\subsection{Traditional Machine Learning Algorithms}

In \cite{ILIAS2021107360}, the authors proposed two approaches for detecting bots in Twitter both at account and at tweet-level. In terms of the account level classification, the authors extracted a large set of features per user, applied feature selection algorithms, sampling techniques for dealing with the imbalanced datasets, and trained shallow machine learning algorithms. 

Similarly, the authors in \cite{KUDUGUNTA2018312} adopted methods for recognizing bots both at account and tweet level. In terms of the account level, the authors extracted a set of features, including number of statuses (number of tweets and retweets), Followers Count, Friends Count, Favorites Count, Listed Count, Default Profile, etc. Next, the authors combined SMOTE with data enhancement via edited nearest neighbors (SMOTENN) and Tomek Links (SMOTOMEK), and trained traditional machine learning algorithms. In terms of the tweet level classification, the authors introduced a deep neural network consisting of LSTM and dense layers. The authors used as input GloVe embeddings of tweets and tweet metadata. Auxiliary output was also used, whose target was also the classification label.

The authors in \cite{ouni2022bots} extracted a set of style-based features, including  the number of punctuation marks, number of hashtags, number of retweets, number of user mentions, number of url links, and many more. They trained the following classification algorithms: Random Forest, Naive Bayes, Logistic Regression, Support Vector Machine with rbf and linear kernel, and Convolution Neural Network.

A wavelet-based approach was introduced by \cite{10.1145/3183506}. Specifically, the authors exploited the discrete wavelet transform and extracted a set of features, namely wavelet magnitude, wavelet phase, wavelet domain score, and so on. After this, a feature selection technique, namely a Correlation-based Feature Subset Selection, was proposed for reducing the dimensionality of the feature vector. Finally, a Random Forest Classifier was trained. 

The authors in \cite{10.1145/2872518.2889302} proposed \textit{Botometer}, which is a publicly available bot detection tool via the website, Python, or REST APIs. The authors use network, user, friends, and temporal features and train a Random Forest Classifier. The only input to this publicly available tool is the screen name.

\subsection{Deep Learning and Transformer-based Approaches}

In \cite{10064110}, the authors exploited three types of LSTMs and four types of features, namely tweet content, tweet metadata, account metadata, and user description. In terms of the tweet content and the user description, the authors exploited the GloVe embeddings. Regarding account metadata, the authors extracted 30 features, while 15 tweet metadata features were also extracted. One limitation of this study is the feature extraction procedure and the usage of GloVe embeddings in conjunction with LSTM models instead of pretrained language models based on transformers, which capture better the context and require significantly less time for fine-tuning.

A similar approach was introduced by \cite{hayawi2022deeprobot}, where the authors utilized GloVe embeddings for representing the user description field as embeddings. Also, the authors extracted several types of features, namely number of followers and friends, number of tweets and retweets posted by the user, length of the name, entropy of the screen name, entropy of the description, etc. They passed the GloVe embeddings of the user description field into LSTM models and concatenated the output of the LSTMs with the rest of the features.

In \cite{ILIAS2021107360}, the authors proposed two approaches for detecting bots in Twitter both at account and at tweet-level. With regards to the tweet-level classification, the authors utilized a deep neural network. Specifically, the authors exploited GloVe embeddings and passed them through BiLSTM layers coupled with an attention mechanism. 

Similarly, the authors in \cite{KUDUGUNTA2018312} adopted methods for recognizing bots both at account and tweet level. In terms of the tweet level classification, the authors introduced a deep neural network consisting of LSTM and dense layers. The authors used as input GloVe embeddings of tweets and tweet metadata. Auxiliary output was also used, whose target was also the classification label.

A different approach was proposed by \cite{najari2022ganbot}, where the authors exploited Generative Adversarial Neural networks (GANs) for recognizing bots in Twitter. Specifically, the authors addressed the limitation of original SeqGAN \cite{10.5555/3298483.3298649} by introducing the GANBOT setting, where a shared LSTM was used between generator and classifer. Similar to \cite{KUDUGUNTA2018312}, the authors used GloVe embeddings as input to LSTMs. GANs were also exploited by \cite{9006873}. Specifically, the authors extracted first a set of features consisting of user meta-data, sentiment, friends, content, network, and timing. Finally, they proposed a new method based on Conditional GANs by introducing the Wasserstein distance with a gradient penalty and a condition-generation method through the modified density peak clustering algorithm. The authors compared their approaches with Random Oversampling, SMOTE, and ADASYN.

In \cite{9385071}, the authors use language models based on transformers, namely BERT and RoBERTa, for recognizing bots in Twitter. The authors concatenated the text embeddings along with additional metadata and trained a neural network consisting of dense layers.

In \cite{10.1007/978-3-031-35995-8_42}, the authors proposed a new method for recognizing bots in Twitter based on image recognition. Specifically, they introduced a new approach for converting the digital DNA sequence into an image consisting of three channels. Finally, the authors applied ResNet and stated that this method achieved competitive results to state-of-the-art approaches. However, the authors created a digital DNA sequence based only on the type of the tweets, while they fine-tuned only one pretrained model.

The authors in \cite{chawla2023hybrid} utilized the concept of digital DNA sequences along with pretrained BERT models. First, the authors trained BERT on sentiment analysis classification tasks. Next, they passed each tweet through a BERT model and predicted the sentiment of each tweet, i.e., positive, negative, and neutral. In this way, they create a digital DNA sequence consisting of characters denoting the sentiment of each tweet.

A deep learning architecture consisting of two branches was proposed by \cite{pr10030439}. Specifically, the authors used word2vec and passed the respective embeddings through CNNs. Next, they extracted features, including number of followers, number of tweets, account age/reputation, etc., and passed them through dense layers. After this, they concatenated the two branches and passed the resulting vector through a dense layer for getting the final prediction.

An active learning approach was adopted by \cite{WU2021106525}. Specifically, the authors extracted a set of features, including metadata-based, interaction-based, content-based, and timing-based. Next, active learning was employed for efficiently expanding the labeled data. Finally, a deep neural network consisting of ResNet, BiGRU, and attention layer was trained.

The study in \cite{9505695} introduced a CNN and BiLSTM-based deep neural network model coupled with an attention mechanism. Specifically, the authors passed the profile, temporal, and activity information to a two-layers stacked BiLSTM, whereas they passed the content
information to a deep CNN. An attention mechanism was exploited at the top of the proposed architecture. Findings showed that the proposed approach outperformed the state-of-the-art approaches.

A CNN-LSTM network was also proposed by \cite{10.1145/3132847.3133050}. Specifically, the authors model the social behaviour per user, i.e., posting and retweeting, and exploit an LSTM network. Next, the authors consider users’ history tweets as temporal text data and use a CNN-LSTM network. Finally, they fuse the representation vectors obtained by the two networks and get the final prediction.

In \cite{10210119}, the authors extract word embeddings, character embeddings, part-of-speech embeddings, and named-entity embeddings. They pass these embeddings through BiLGRU \cite{9297350} and get the final prediction. Findings stated that the proposed model performed better or comparably to state-of-the-art Twitter bot detection models.

In \cite{9014365}, the authors used GloVe embeddings as input to a 3-layer BiLSTM neural network for distinguishing Twitter bots from human accounts.

\subsection{Graph-based Approaches}

Recently, graph convolutional networks have been also used for identifying bots in Twitter. Specifically, the study in \cite{10.1145/3487351.3488336} created an heterogeneous graph and employed Relational Graph Convolutional Networks. The authors represented each user as a vector consisting of user description, tweets, numerical and categorical properties. To be more precise, the authors employed RoBERTa to get a representation vector of the user description. Similarly, they used RoBERTa and averaged the representations of all the tweets posted by the user into one single vector. In terms of the numerical properties, the authors created a vector consisting of the number of friends, followers, favourites, etc. Regarding the categorical properties, the resulting vector was composed of binary variables, i.e., if the user has a profile image, if the user is verified, etc. However, this method cannot incorporate the intrinsic heterogeneity of relation. Therefore, the authors in \cite{Feng_Tan_Li_Luo_2022} introduced propose a novel Twitter bot detection framework that
is graph-based and heterogeneity-aware. Specifically, the authors encoded each user adopting the procedure followed in \cite{10.1145/3487351.3488336}. Finally, the authors exploited relational graph transformers and semantic attention networks.

The authors in \cite{lei2023bic} proposed a text-graph interaction module along with a semantic consistency module. Specifically, the authors exploited the description and tweets and passed them through RoBERTa for getting the textual representation (embeddings). For graph-based approach, they adopted the method proposed by \cite{10.1145/3487351.3488336}. Next, they introduced a method for capturing the interactions between the text and graph modality. After that, a semantic consistency detection module was proposed, which exploits the attention weights obtained by the RoBERTa model. Finally, the respective vectors were concatenated and were passed through a dense layer for the final classification.

Multi-view graph attention networks were exploited in \cite{app12168117}. The authors used different datasets for training and evaluating their proposed approaches under a transfer learning scenario. Profile features were also utilized, including number of followers/friends, age in days, and many more.

Motivated by the fact that existing methods almost ignore the differences in bot behaviors in multiple domains, the study in \cite{9892366} introduces a domain-aware approach for recognizing bots. Specifically, multi-relational graphs were exploited coupled with a user representation learning module, consisting of a series of graph embedding layers and semantic attention layers. Finally, domain-aware classifiers were exploited for detecting bots. Additionally, according to the authors this is the first study employing a federated learning framework. 

The authors in \cite{10.1145/3308560.3316504} introduced a Graph Convolutional Neural Network for exploiting the characteristics of the accounts' neighbours and identifying bots. Each user is represented by a node consisting of the following feature set: age, favourites\_count, statuses\_count, account length name, followers\_count, and friends\_count. Results showed that the proposed approach outperformed the state-of-the-art ones.

\subsection{Unsupervised Learning} \label{unsupervised_learning_related_work}

Cresci et al. \cite{7876716} introduced digital DNA for recognizing bots in Twitter. Specifically, the authors set the tweets of each user into chronological user and created a digital DNA sequence according to the type and content of each tweet. In terms of the type of the tweets, the resulting DNA sequence was composed of A, C, and T, where a "C" indicates a reply, a "T" denotes a retweet, and an "A" denotes a simple tweet. With regards to the content of the tweet, the authors examined whether the tweet contained hashtags, URLs, mentions, media, etc. Finally, the authors calculated similarities between these sequences using the longest common subsequence (LCS) similarity and clustered users based on their similarity scores. 

In \cite{10020363}, the authors introduce MulBot, which constitutes an unsupervised bot detector based on multivariate time series. Specifically, the proposed approach consists of the following steps: multidimensional
temporal features extracted from user timelines, dimensionality reduction using an autoencoder, extraction of statistical global features (optional), concatenation of global features with vectorial features in output from the encoder (optional), and clustering algoritm. Results showed that the proposed approach yielded satisfactory results in both the binary task and the multiclass classification one. 

The bot detection task was considered as an anomaly detection task in \cite{MILLER201464}. The authors utilized 95 one-gram features from tweet text along with user features and modified two stream clustering algorithms, namely StreamKM++ and DenStream.

\subsection{Reinforcement Learning}

 A reinforcement learning approach was proposed by \cite{10.1145/3572403} for searching the GNN architecture. In this way, the most suitable multi-hop neighborhood and the number of
layers in the GNN architecture are found. An Heterogeneous Information Network was also exploited for modelling the entities and relationships in the social networks. Finally, the authors exploited self-supervised learning approaches. Also, the authors in \cite{lingam2019adaptive} introduced a deep Q-network architecture by incorporating a Deep Q-Learning (DQL) model. The authors extracted tweet-based, user profile-based, and social graph-based features. In terms of the tweet-based, the authors utilized syntax, semantic, and temporal behaviour features. Regarding the user profile features, the authors employed features pertaining to user behaviour (Posting tweets in several languages, URL/hashtag/mention ratio) and user interactions (Number of active days, Number of retweeted tweets). Finally, they defined social graph-based attributes, such as clustering coefficient, closeness centrality, betweenness centrality and pagerank centrality for each user.

The authors in \cite{9093747} extracted URL-based features, including URL redirection, frequency of shared URLs, and spam content in URL, and designed a learning automata based model. Specifically, the authors designed a trust computation model, which contains two parameters, namely direct trust and indirect trust. Findings suggested that the proposed approach improved the existing ones.

\subsection{Related Work Review Findings}

In spite of the rise of deep learning algorithms, existing research initiatives extract still a large number of features and train traditional machine learning algorithms. However, feature extraction constitutes a tedious procedure demanding domain expertise. Thus, the optimal set of features may not be found. Additionally, existing research initiatives still exploit GloVe embeddings and train LSTMs instead of employing language models based on transformers, which achieve state-of-the-art results across a large number of domains. In addition, methods employing BERT cannot be employed for languages other than the english one. Recently, there have been introduced graph-based approaches. However, these methods still extract a large number of features per user for creating a feature vector per node-user. At the same time, large graphs are created, which require access to computational resources and need a lot of time for training.

Therefore, our proposed work differs significantly from the aforementioned research works, since we \textit{(1)} exploit only the user description field and create images according to the type and content of the tweets posted by each user, \textit{(2)} create images consisting of three channels and train several pretrained models comparing their performances, \textit{(3)} employ TwHIN-BERT which is a new multi-lingual Tweet language model that is trained on 7 billion Tweets from over 100 distinct languages, \textit{(4)} introduce multimodal models, which take as input the user description field and the 3d images.

\section{Datasets}

 \subsection{Cresci'17 Dataset}

We use the dataset introduced in \cite{10.1145/3041021.3055135} for conducting our experiments. Specifically, this dataset consists of genuine accounts and some sets of social and traditional spambots. In the present study, we exploit the set of genuine accounts and the set of social spambots \#1. For collecting genuine users, the authors adopted the methodology proposed in \cite{10.1145/3041021.3051155}. Specifically, the authors contacted with some accounts and asked them a question. Then, the authors examined the users, who replied to the question and verified 3,474 accounts. In terms of the social spambots users, the authors used a group of social bots that was discovered on Twitter during the last Mayoral election in Rome, in 2014. Specifically, this set of social spambots corresponds to retweeters of an Italian political candidate. As mentioned in Section~\ref{unsupervised_learning_related_work}, the authors presented a figure illustrating the LCS curves of humans and bots. Findings showed that the LCS of bots were long, even when the number of account increases. A sudden drop
in LCS length of bots is observed when the number of accounts gets close to the group size. On the contrary, the LCS curve of humans indicates little to no similarity. Also, in \cite{10.1145/3041021.3055135}, the authors illustrate the cumulative distribution function (CDF) of join date and number of followers. Results indicate that social spambots have anomalous distributions. In our experiments, we keep users having available the user description field in their profile. To address the issue of the imbalanced dataset, we create a dataset consisting of 943 real users and 943 social spambots. This technique, which addresses the issue of imbalanced dataset by downsampling the set of real users, has been adopted by previous studies \cite{10.1007/978-3-031-35995-8_42,10020363}.

 \subsection{TwiBot-20 Dataset}

We use the TwiBot-20 dataset to conduct our experiments \cite{10.1145/3459637.3482019}. Contrary to other datasets which include a specific type of users, this dataset consists of diversified bots that co-exist on the real-world Twittersphere. Users of this dataset have an interest in four domains, including politics, business, entertainment, and sports. Thus, this dataset constitutes a challenging task, since it includes multiple types of bots. For proving the high quality of annotations, the authors illustrate the CDF of account reputation (number of friends and followers), user tweet counts, and screen name likelihood. CDF plots show that genuine accounts present higher reputation scores than bots in the TwiBot-20 dataset. In terms of user tweet counts, CDF plots indicate that bots generate fewer tweets than humans, in order to avoid the traditional detection methods. Finally, findings of CDF plot regarding the screen name likelihood show that  bot users in TwiBot-20 do have slightly lower screen name likelihood. Also, the authors in \cite{10.1145/3459637.3482019} conduct a user diversity analysis and study the distribution of profile locations and user interests. Findings of a geographic analysis study demonstrate that India, United States, Europe, and Africa are represented in the TwiBot-20 dataset. Finally, the authors examine the most frequent hashtags, present their analysis through a bar plot, and show that twitter users in TwiBot-20 are proved to be diversified in interest domains.
TwiBot-20 dataset includes 229,573 users. However, the authors have labelled 11,826 users. This dataset includes the 200 recent tweets per user.
The authors in \cite{10.1145/3459637.3482019} have divided TwiBot-20 dataset into a train, validation, and test set. We remove users who have not posted any tweets and have not added a description field in their profile. Table~\ref{twibot20_characteristics} presents the distribution of genuine accounts and bots into each set. 

\begin{table}[!htb]
    \centering
    \caption{TwiBot-20 dataset statistics.}
    
    \begin{tabular}{|>{\columncolor[gray]{0.8}}c|c|c|}
    \toprule
    \rowcolor{gray}
         & genuine users & bots \\ \hline
      train set & 3,262 & 3,911 \\ \hline
      development set   & 959 & 1,101\\ \hline
      test set & 494 & 533\\
      \bottomrule
    \end{tabular}
    \label{twibot20_characteristics}
\end{table}

\section{Unimodal Models Utilizing only Images for Detecting Bots in Twitter} \label{unimodal_images}

We adopt the methodology introduced by \cite{7436643,7876716} for constructing a DNA sequence. Inspired by the biological DNA sequence consisting of A (adenine), C (cytosine), G (guanine) and T (thymine), the authors in \cite{7436643,7876716} introduce the digital DNA for creating a DNA sequence based either on type of the tweet or the content of the tweet. In this study, we create two digital DNA sequences based on both the type and content of the tweet. First, we set all tweets into a chronological order. Regarding the first approach, we create a sequence consisting of A, T, and C. A "T" denotes a retweet, a "C" indicates a reply, while an "A" denotes a tweet. In terms of the second approach, which is based on the content of the tweet, we create a digital DNA sequence consisting of N, U, H, M, and X. Specifically, a "N" indicates that the tweet contains no entities (plain text), a "U" denotes that the tweet contains one or more ULRs, a "H" means that the tweet contains one or more hashtags, a "M" indicates that the tweet contains one or more mentions, and a "X" indicates that the tweet contains entities of mixed types.

After having created a digital DNA sequence per user, we adopt the method introduced in \cite{10.1007/978-3-031-35995-8_42} for transforming the DNA sequence into an image consisting of three channels. First, we check whether the length of the longest DNA sequence is a perfect square. If it is a perfect square, then we define the image size as the square root of the length of the longest DNA sequence. Otherwise, we consider the perfect square closest to and larger than the maximum length. In this way, all strings can be converted to images of equal sizes. Next, we assign to each symbol of the DNA sequence, i.e., A, C, and T, a "color" for creating the image. The image is colored pixel by pixel based on the coors assigned to the
correspondent symbol. The same methodology is adopted in terms of the digital DNA sequence based on the content of the tweets, which consists of N, U, H, M, and X. The resulting image is a grayscale image in a (1, H, W) format. Similar to \cite{10.1007/978-3-031-35995-8_42}, we convert the images into a (3, H, W) format utilizing the grayscale transformation\footnote{https://pytorch.org/vision/main/generated/torchvision.transforms.Grayscale. html}.

Each image is resized to $256 \times 256$ pixels. Next, we fine-tune the following pretrained models: \textbf{GoogLeNet (Inception v1)} \cite{7298594}, \textbf{ResNet50} \cite{7780459}, \textbf{WideResNet-50-2} \cite{zagoruyko2016wide}, \textbf{AlexNet} \cite{krizhevsky2014one}, \textbf{SqueezeNet1\_0} \cite{iandola2016squeezenet}, \textbf{DenseNet-201} \cite{8099726}, \textbf{MobileNetV2}\cite{8578572}, \textbf{ResNeXt-50 32$\times$4d} \cite{8100117}, \textbf{VGG16} \cite{simonyan2014very}, and \textbf{EfficientNet-B4}\footnote{We experimented with EfficientNet-B0 to B7, but EfficientNet-B4 was the best performing model.} \cite{pmlr-v97-tan19a}.

\subsection{Experiments}

\subsubsection{Experimental Setup} \label{section_experimental_setup}

In terms of the Cresci'17 dataset, we divide the dataset into a train, validation, and test set (80\%-10\%-10\%). Regarding the TwiBot-20 dataset, the train, validation, and test sets are provided by the authors \cite{10.1145/3459637.3482019}. We use \textit{EarlyStopping} and stop training if the validation loss has stopped decreasing for 6 consecutive epochs. We use Adam optimizer with a learning rate of 1e-5 and exploit \textit{ReduceLROnPlateau}, where we reduce the learning rate by a factor of 0.1 if the validation loss has not presented an improvement after 3 consecutive epochs. We minimize the cross-entropy loss function. In terms of the TwiBot-20 dataset, we apply class-weights to the loss function to deal with the data imbalance.  We train the proposed approaches for a maximum of 30 epochs. All models have been created using the PyTorch library \cite{NEURIPS2019_9015}. All experiments are conducted on a single Tesla T4 GPU.

\subsubsection{Evaluation Metrics} \label{section_evaluation_metrics}

We use Precision, Recall, F1-score, Accuracy, and Specificity for evaluating the results of the proposed approaches. These metrics have been computed by considering the class of bots as the positive one (label=1). Results are obtained over five runs using different random seeds
reporting the average and the standard deviation.

\subsubsection{Results} \label{results_unimodal_images}

The results of our proposed approaches are reported in Tables \ref{unimodal_models_image_type}-\ref{unimodal_models_image_content_twibot}. Specifically, Tables~\ref{unimodal_models_image_type} and \ref{unimodal_models_image_content} report the results on the Cresci'17 dataset, while Tables~\ref{unimodal_models_image_type_twibot} and \ref{unimodal_models_image_content_twibot} report the results on the TwiBot-20 dataset.

\paragraph{Cresci'17 Dataset} The results of our proposed approaches are reported in Tables \ref{unimodal_models_image_type} and \ref{unimodal_models_image_content}.

As one can easily observe in Table~\ref{unimodal_models_image_type}, VGG16 constitutes the best performing model outperforming the other pretrained models in Recall, F1-score, and Accuracy. It is worth noting that there are models surpassing VGG16 or obtaining equal performance in Precision and Specificity. However, VGG16 outperforms these models in F1-score, which constitutes the weighted average of Precision and Recall. Additionally, F1-score is a more important metric than Specificity, since high Specificity and low F1-score indicates that some social spambots are falsely detected as real accounts. VGG16 outperforms the other models in Recall by 0.44-3.48\%, in F1-score by 0.22-2.06\%, and in Accuracy by 0.19-2.12\%. WideResNet-50-2 constitutes the second best performing model attaining an Accuracy of 99.58\% and a F1-score of 99.57\%. Specifically, it surpasses the other models, except for VGG16, in Accuracy by 0.64-1.91\% and in F1-score by 0.59-1.84\%. F1-score ranging from 98.17\% to 98.98\% is obtained by the rest of the models, except for EfficientNet-B4. Similarly, Accuracy ranging from 98.20\% to 98.94\% is attained by the rest of the models, except for EfficientNet-B4. The highest Precision and Specificity scores are obtained by GoogLeNet and ResNet50 and are equal to 1.00. EfficientNet-B4 obtains the worst evaluation results reaching Accuracy and F1-score up to 97.67\% and 97.73\% respectively. 

As one can easily observe in Table~\ref{unimodal_models_image_content}, VGG16 constitutes our best performing model obtaining an Accuracy of 99.89\%, a Precision of 1.00\%, a Recall of 99.78\%, a F1-score of 99.89\%, and a Specificity of 1.00\%. Specifically, it outperforms the other models in Recall by 0.44-2.14\%, in F1-score by 0.52-2.87\%, and in Accuracy by 0.52-2.75\%, while it achieves equal Precision and Specificity scores with GoogLeNet, ResNet50, and AlexNet. Additionally, GoogLeNet, WideResNet-50-2, AlexNet, and MobileNetV2 achieve Accuracy and F1-scores over 99.00\%. AlexNet constitutes the second best performing model obtaining an Accuracy of 99.37\% and a F1-score of 99.37\%. EfficientNet-B4 achieves the worst evaluation results reaching Accuracy and F1-score up to 97.14\% and 97.02\% respectively.   

\begin{table}[!htb]
\tiny
\centering
\caption{Performance comparison among proposed models on the Cresci'17 dataset (using only images based on the type of the tweet). Reported values are mean $\pm$ standard deviation.  Results are averaged across five runs. Best results per evaluation metric are in bold.}
\begin{tabular}{lccccc}
\toprule
\multicolumn{1}{l}{}&\multicolumn{5}{c}{\textbf{Evaluation metrics}}\\
\cline{2-6} 
\multicolumn{1}{l}{\textbf{Architecture}}&\textbf{Precision}&\textbf{Recall}&\textbf{F1-score}&\textbf{Accuracy}&\textbf{Specificity}\\ \midrule 
\multicolumn{6}{>{\columncolor[gray]{.8}}l}{\textbf{digital DNA (type of tweets)}} \\ \hline
\textit{\tiny{GoogLeNet (Inception v1)}} & \textbf{100.00} & 97.63 & 98.79 & 98.73 & \textbf{100.00} \\
& $\pm$0.00 & $\pm$1.39 & $\pm$0.72 & $\pm$0.79 & $\pm$0.00 \\ \hline
\textit{\tiny{ResNet50}} & \textbf{100.00} & 97.83 & 98.90 & 98.94 & \textbf{100.00} \\
& $\pm$0.00 & $\pm$1.00 & $\pm$0.52 & $\pm$0.47 & $\pm$0.00 \\ \hline
\textit{\tiny{WideResNet-50-2}} & 99.78 & 99.35 & 99.57 & 99.58 & 99.79 \\
& $\pm$0.44 & $\pm$1.29 & $\pm$0.87 & $\pm$0.85 & $\pm$0.42 \\ \hline
\textit{\tiny{AlexNet}} & 99.59 & 97.12 & 98.34 & 98.31 & 99.55 \\
& $\pm$0.81 & $\pm$1.23 & $\pm$0.88 & $\pm$0.91 & $\pm$0.91 \\ \hline
\textit{\tiny{SqueezeNet1\_0}} & 99.17 & 97.23 & 98.17 & 98.20 & 99.14 \\
& $\pm$1.02 & $\pm$1.41 & $\pm$0.55 & $\pm$0.54 & $\pm$1.05 \\ \hline
\textit{\tiny{DenseNet-201}} & 99.37 & 98.59 & 98.98 & 98.94 & 99.35 \\
& $\pm$0.51 & $\pm$0.98 & $\pm$0.55 & $\pm$0.58 & $\pm$0.53 \\ \hline
\textit{\tiny{MobileNetV2}} & 99.58 & 97.70 & 98.63 & 98.62 & 99.57 \\
& $\pm$0.52 & $\pm$1.28 & $\pm$0.76 & $\pm$0.72 & $\pm$0.53\\ \hline
\textit{\tiny{ResNeXt-50 32 $\times$ 4d}} & 99.79 & 97.59 & 98.67 & 98.62 & 99.78 \\
& $\pm$0.43 & $\pm$1.64 & $\pm$1.01 & $\pm$1.04 & $\pm$0.44 \\ \hline
\textit{\tiny{VGG16}} & 99.78 & 99.55 & 99.67 & 99.68 & 99.80 \\
& $\pm$0.44 & $\pm$0.54 & $\pm$0.44 & $\pm$0.42 & $\pm$0.41 \\ \hline
\textit{\tiny{EfficientNet-B4}} & 99.20 & 96.31 & 97.73 & 97.67 & 99.08 \\
& $\pm$1.14 & $\pm$0.60 & $\pm$0.47 & $\pm$0.54 & $\pm$1.35 \\ 
\bottomrule
\end{tabular}
\label{unimodal_models_image_type}
\end{table}

\begin{table}[!htb]
\tiny
\centering
\caption{Performance comparison among proposed models on the Cresci'17 dataset (using only images based on the content of the tweet). Reported values are mean $\pm$ standard deviation. Results are averaged across five runs. Best results per evaluation metric are in bold.}
\begin{tabular}{lccccc}
\toprule
\multicolumn{1}{l}{}&\multicolumn{5}{c}{\textbf{Evaluation metrics}}\\
\cline{2-6} 
\multicolumn{1}{l}{\textbf{Architecture}}&\textbf{Precision}&\textbf{Recall}&\textbf{F1-score}&\textbf{Accuracy}&\textbf{Specificity}\\ \midrule 
\multicolumn{6}{>{\columncolor[gray]{.8}}l}{\textbf{digital DNA (content of tweets)}} \\ \hline
\textit{\tiny{GoogLeNet (Inception v1)}} & \textbf{100.00} & 98.09 & 99.03 & 99.05 & \textbf{100.00} \\
& $\pm$0.00 & $\pm$1.27 & $\pm$0.65 & $\pm$0.62 & $\pm$0.00 \\ \hline
\textit{\tiny{ResNet50}} & \textbf{100.00} & 97.81 & 98.89 & 98.84 & \textbf{100.00} \\
& $\pm$0.00 & $\pm$1.13 & $\pm$0.58 & $\pm$0.62 & $\pm$0.00 \\ \hline
\textit{\tiny{WideResNet-50-2}} & 98.93 & 99.34 & 99.14 & 99.15 & 98.95 \\
& $\pm$0.71 & $\pm$0.89 & $\pm$0.69 & $\pm$0.63 & $\pm$0.64 \\ \hline
\textit{\tiny{AlexNet}} & \textbf{100.00} & 98.75 & 99.37 & 99.37 & \textbf{100.00} \\
& $\pm$0.00 & $\pm$1.20 & $\pm$0.61 & $\pm$0.62 & $\pm$0.00 \\ \hline
\textit{\tiny{SqueezeNet1\_0}} & 99.36 & 97.64 & 98.49 & 98.52 & 99.36 \\
& $\pm$0.53 & $\pm$1.06 & $\pm$0.64 & $\pm$0.62 & $\pm$0.52 \\ \hline
\textit{\tiny{DenseNet-201}} & 99.59 & 97.90 & 98.73 & 98.73 & 99.56 \\
& $\pm$0.49 & $\pm$1.37 & $\pm$0.58 & $\pm$0.54 & $\pm$0.54 \\ \hline
\textit{\tiny{MobileNetV2}} & 99.59 & 98.98 & 99.28 & 99.26 & 99.56 \\
& $\pm$0.50 & $\pm$1.27 & $\pm$0.74 & $\pm$0.79 & $\pm$0.54\\ \hline
\textit{\tiny{ResNeXt-50 32 $\times$ 4d}} & 99.35 & 98.43 & 98.88 & 98.94 & 99.38 \\
& $\pm$0.53 & $\pm$1.31 & $\pm$0.52 & $\pm$0.47 & $\pm$0.51 \\ \hline
\textit{\tiny{VGG16}} & \textbf{100.00} & \textbf{99.78} & \textbf{99.89} & \textbf{99.89} & \textbf{100.00} \\
& $\pm$0.00 & $\pm$0.43 & $\pm$0.22 & $\pm$0.21 & $\pm$0.00 \\ \hline
\textit{\tiny{EfficientNet-B4}} & 96.13 & 97.98 & 97.02 & 97.14 & 96.36 \\
& $\pm$2.26 & $\pm$2.01 & $\pm$1.38 & $\pm$1.28 & $\pm$2.14 \\ 
\bottomrule
\end{tabular}
\label{unimodal_models_image_content}
\end{table}

\paragraph{TwiBot-20 Dataset}

 The results of our proposed approaches are reported in Tables \ref{unimodal_models_image_type_twibot} and \ref{unimodal_models_image_content_twibot}.

As one can easily observe in Table~\ref{unimodal_models_image_type_twibot}, AlexNet constitutes the best performing model in terms of F1-score and Accuracy. Specifically, it outperforms the other models in F1-score by 0.48-1.34\% and in Accuracy by 0.06-0.78\%. Although other models outperform AlexNet in Precision and Recall, F1-score is a more important metric, since it is a weighted average of precision and recall. DenseNet-201 and MobileNetV2 achieve almost equal Accuracy scores accounting for 66.35\% and 66.34\% respectively. The second highest F1-score is achieved by SqueezeNet1\_0 and is equal to 66.63\%. GoogLeNet obtains the worst performance reaching Accuracy and F1-score up to 65.63\% and 65.71\% respectively. 

As one can easily observe in Table~\ref{unimodal_models_image_content_twibot}, AlexNet is the best performing model outperforming the rest of the models in Precision by 2.11-3.16\%, in F1-score by 0.78-3.06\%, in Accuracy by 1.60-2.50\%, and in Specificity by 1.13-4.58\%. GoogLeNet, SqueezeNet1\_0, MobileNetV2, VGG16, and EfficientNet-B4 obtain almost equal Accuracy scores ranging from 65.45\% to 65.59\%, with MobileNetV2 outperforming these models in terms of both Accuracy and F1-score. ResNet50 obtains the worst performance reaching Accuracy and F1-score up to 64.69\% and 63.74\% respectively.

\begin{table}[!htb]
\tiny
\centering
\caption{Performance comparison among proposed models on the TwiBot-20 dataset (using only images based on the type of the tweet). Reported values are mean $\pm$ standard deviation.  Results are averaged across five runs. Best results per evaluation metric are in bold.}

\begin{tabular}{lccccc}
\toprule
\multicolumn{1}{l}{}&\multicolumn{5}{c}{\textbf{Evaluation metrics}}\\
\cline{2-6} 
\multicolumn{1}{l}{\textbf{Architecture}}&\textbf{Precision}&\textbf{Recall}&\textbf{F1-score}&\textbf{Accuracy}&\textbf{Specificity}\\ \midrule 
\multicolumn{6}{>{\columncolor[gray]{.8}}l}{\textbf{digital DNA (type of tweets)}} \\ \hline
\textit{\tiny{GoogLeNet (Inception v1)}} & 68.12 & 63.53 & 65.71 & 65.63 & 67.89 \\
& $\pm$0.61 & $\pm$2.58 & $\pm$1.29 & $\pm$0.67 & $\pm$1.84 \\ \hline
\textit{\tiny{ResNet50}} & 67.90 & 65.29 & 66.46 & 65.90 & 66.56 \\
& $\pm$1.44 & $\pm$4.30 & $\pm$1.90 & $\pm$0.94 & $\pm$3.98 \\ \hline
\textit{\tiny{WideResNet-50-2}} & 67.53 & 65.63 & 66.52 & 65.76 & 65.91 \\
& $\pm$0.73 & $\pm$2.85 & $\pm$1.30 & $\pm$0.63 & $\pm$2.31 \\ \hline
\textit{\tiny{AlexNet}} & 68.33 & 65.89 & \textbf{67.05} & \textbf{66.41} & 66.96 \\
& $\pm$1.26 & $\pm$2.05 & $\pm$0.72 & $\pm$0.62 & $\pm$2.79 \\ \hline
\textit{\tiny{SqueezeNet1\_0}} & 67.52 & \textbf{66.19} & 66.63 & 65.74 & 65.26 \\
& $\pm$2.13 & $\pm$5.54 & $\pm$1.72 & $\pm$0.26 & $\pm$6.27 \\ \hline 
\textit{\tiny{DenseNet-201}} & 68.86 & 64.32 & 66.45 & 66.35 & 68.54 \\
& $\pm$1.03 & $\pm$3.10 & $\pm$1.31 & $\pm$0.52 & $\pm$2.83 \\ \hline
\textit{\tiny{MobileNetV2}} & \textbf{69.04} & 64.09 & 66.44 & 66.34 & \textbf{68.95} \\
& $\pm$0.79 & $\pm$2.26 & $\pm$0.91 & $\pm$0.37 & $\pm$2.20 \\ \hline
\textit{\tiny{ResNeXt-50 32 $\times$ 4d}} & 68.24 & 64.95 & 66.49 & 66.08 & 67.29 \\
& $\pm$1.15 & $\pm$3.13 & $\pm$1.22 & $\pm$0.43 & $\pm$3.14 \\ \hline
\textit{\tiny{VGG16}} & 67.85 & 65.78 & 66.57 & 65.86 & 65.95 \\
& $\pm$2.35 & $\pm$5.69 & $\pm$1.81 & $\pm$0.67 & $\pm$6.68 \\ \hline
\textit{\tiny{EfficientNet-B4}} & 67.87 & 64.65 & 66.11 & 65.69 & 66.80 \\
& $\pm$1.44 & $\pm$4.17 & $\pm$1.54 & $\pm$0.36 & $\pm$4.25 \\ 
\bottomrule
\end{tabular}
\label{unimodal_models_image_type_twibot}
\end{table}

\begin{table}[!htb]
\tiny
\centering
\caption{Performance comparison among proposed models on the TwiBot-20 dataset (using only images based on the content of the tweet). Reported values are mean $\pm$ standard deviation.  Results are averaged across five runs. Best results per evaluation metric are in bold.}

\begin{tabular}{lccccc}
\toprule
\multicolumn{1}{l}{}&\multicolumn{5}{c}{\textbf{Evaluation metrics}}\\
\cline{2-6} 
\multicolumn{1}{l}{\textbf{Architecture}}&\textbf{Precision}&\textbf{Recall}&\textbf{F1-score}&\textbf{Accuracy}&\textbf{Specificity}\\ \midrule 
\multicolumn{6}{>{\columncolor[gray]{.8}}l}{\textbf{digital DNA (content of tweets)}} \\ \hline
\textit{\tiny{GoogLeNet (Inception v1)}} & 67.53 & \textbf{64.54} & 65.97 & 65.47 & 66.48 \\
& $\pm$0.78 & $\pm$2.12 & $\pm$0.94 & $\pm$0.50 & $\pm$2.07 \\ \hline
\textit{\tiny{ResNet50}} & 68.25 & 59.89 & 63.74 & 64.69 & 69.88 \\
& $\pm$0.91 & $\pm$2.86 & $\pm$1.39 & $\pm$0.52 & $\pm$2.54 \\ \hline
\textit{\tiny{WideResNet-50-2}} & 67.57 & 62.55 & 64.78 & 64.83 & 67.29 \\
& $\pm$2.19 & $\pm$4.80 & $\pm$1.85 & $\pm$0.81 & $\pm$5.59 \\ \hline
\textit{\tiny{AlexNet}} & \textbf{70.36} & 63.64 & \textbf{66.80} & \textbf{67.19} & \textbf{71.01} \\
& $\pm$1.11 & $\pm$1.53 & $\pm$0.37 & $\pm$0.28 & $\pm$2.20 \\  \hline
\textit{\tiny{SqueezeNet1\_0}} & 68.08 & 63.38 & 65.52 & 65.49 & 67.77 \\
& $\pm$1.36 & $\pm$4.16 & $\pm$1.66 & $\pm$0.31 & $\pm$4.01 \\ \hline
\textit{\tiny{DenseNet-201}} & 67.20 & 63.60 & 65.31 & 64.97 & 66.43 \\
& $\pm$1.02 & $\pm$2.39 & $\pm$0.96 & $\pm$0.50 & $\pm$2.70 \\ \hline
\textit{\tiny{MobileNetV2}} & 67.60 & 64.47 & 66.02 & 65.59 & 66.80 \\
& $\pm$0.46 & $\pm$2.38 & $\pm$1.14 & $\pm$0.52 & $\pm$1.71 \\ \hline
\textit{\tiny{ResNeXt-50 32 $\times$ 4d}} & 67.84 & 61.65 & 64.40 & 64.77 & 68.14 \\
& $\pm$2.00 & $\pm$5.19 & $\pm$1.77 & $\pm$0.40 & $\pm$5.86 \\ \hline
\textit{\tiny{VGG16}} & 68.13 & 63.19 & 65.41 & 65.45 & 67.89 \\
& $\pm$1.61 & $\pm$4.95 & $\pm$2.06 & $\pm$0.64 & $\pm$4.74 \\ \hline
\textit{\tiny{EfficientNet-B4}} & 67.56 & 64.50 & 65.96 & 65.47 & 66.52 \\
& $\pm$0.90 & $\pm$2.09 & $\pm$0.66 & $\pm$0.13 & $\pm$2.46 \\ 
\bottomrule
\end{tabular}
\label{unimodal_models_image_content_twibot}
\end{table}

\section{Our Proposed Multimodal Models for Detecting Bots in Twitter} \label{proposed_multimodal}

\begin{figure*}[!htb]
\centering
\begin{subfigure}[t]{1\columnwidth}
\includegraphics[width=1.1\textwidth]{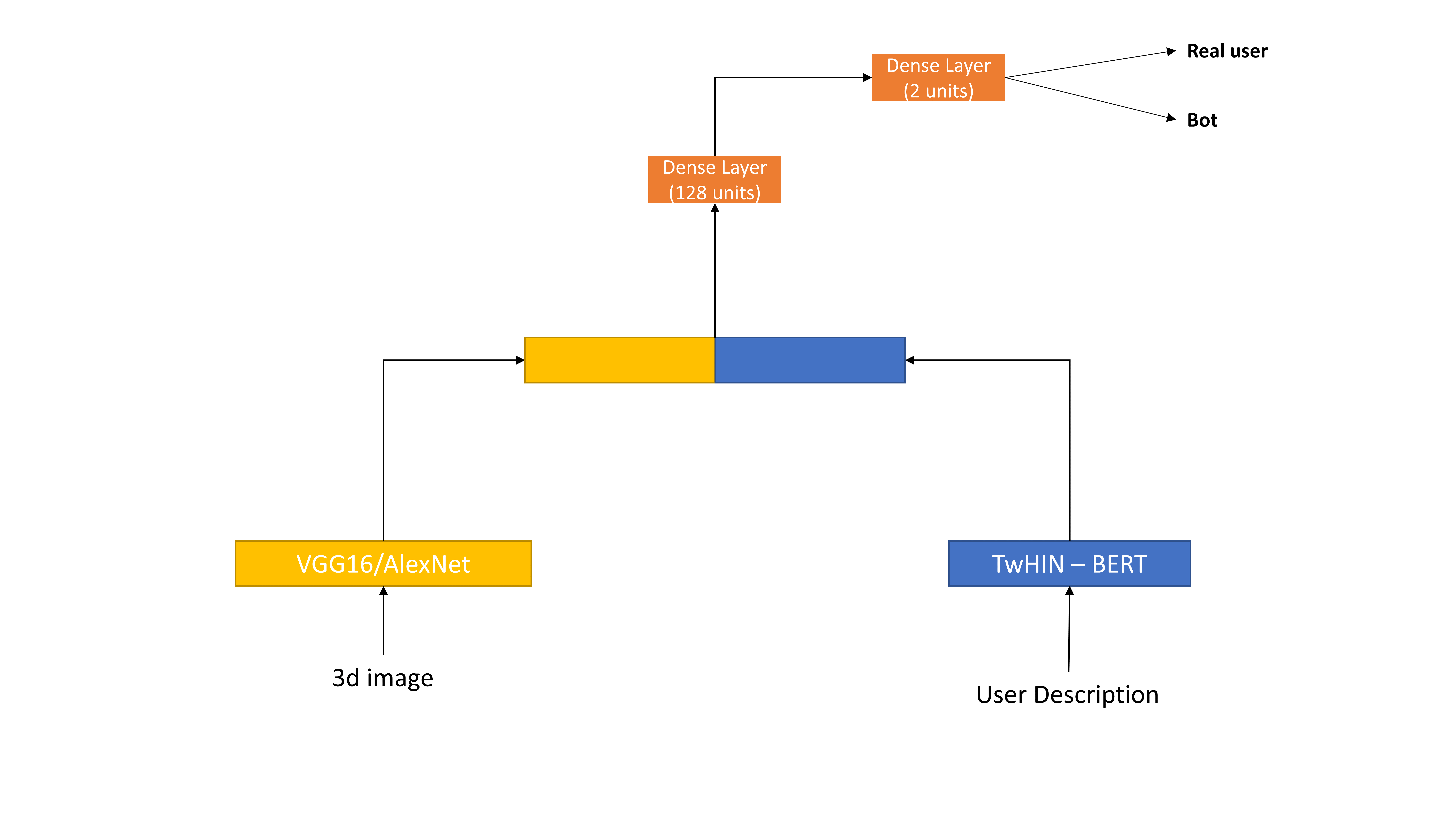}
\caption{Concatenation}
\label{concatenation}
\end{subfigure}
\begin{subfigure}[t]{1\columnwidth}
\centering
\includegraphics[width=1.3\textwidth]{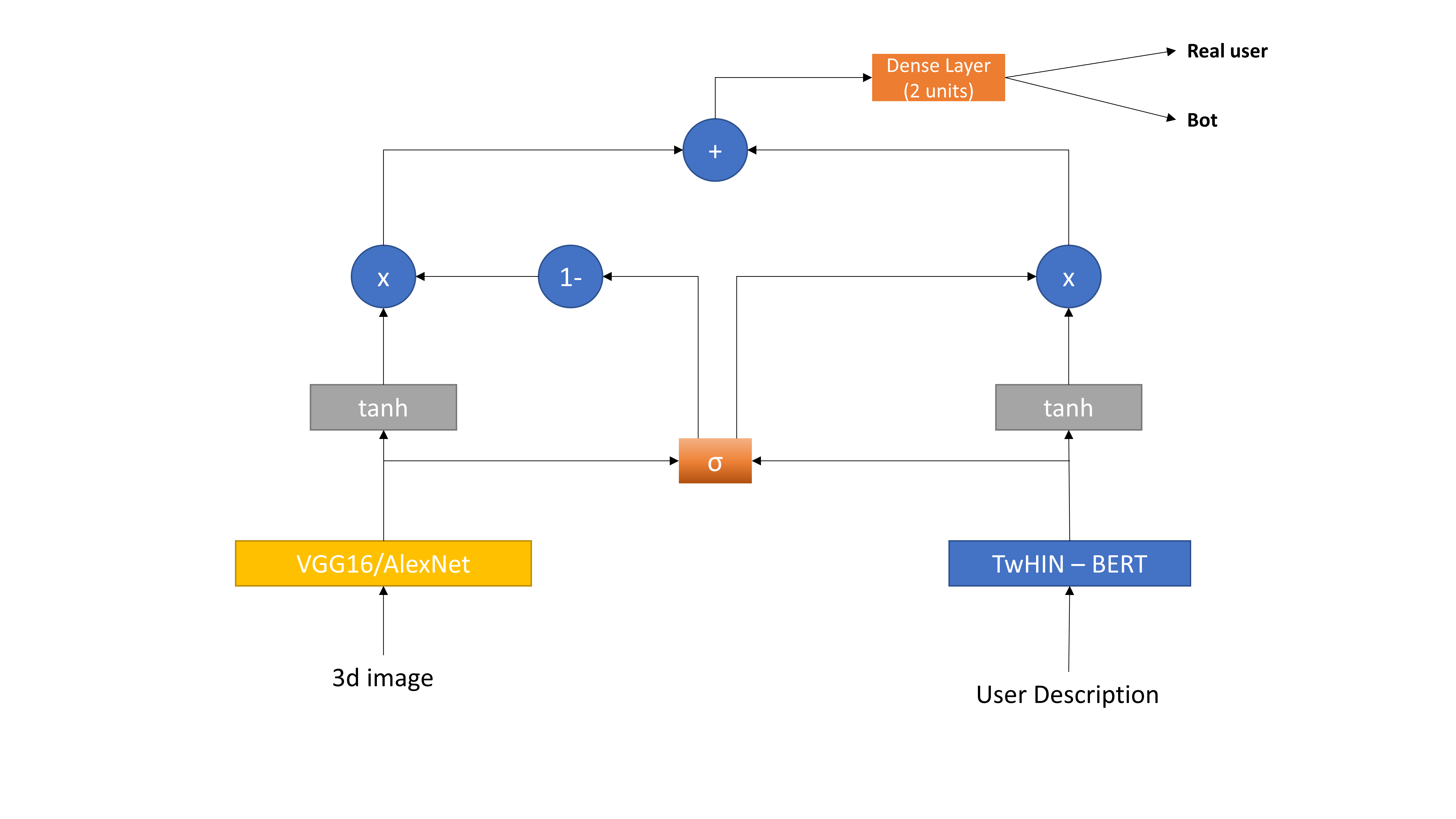}
\caption{Gated Multimodal Unit}
\label{gated_mu}
\end{subfigure}

\begin{subfigure}[t]{1.4\columnwidth}
\includegraphics[width=1\textwidth]{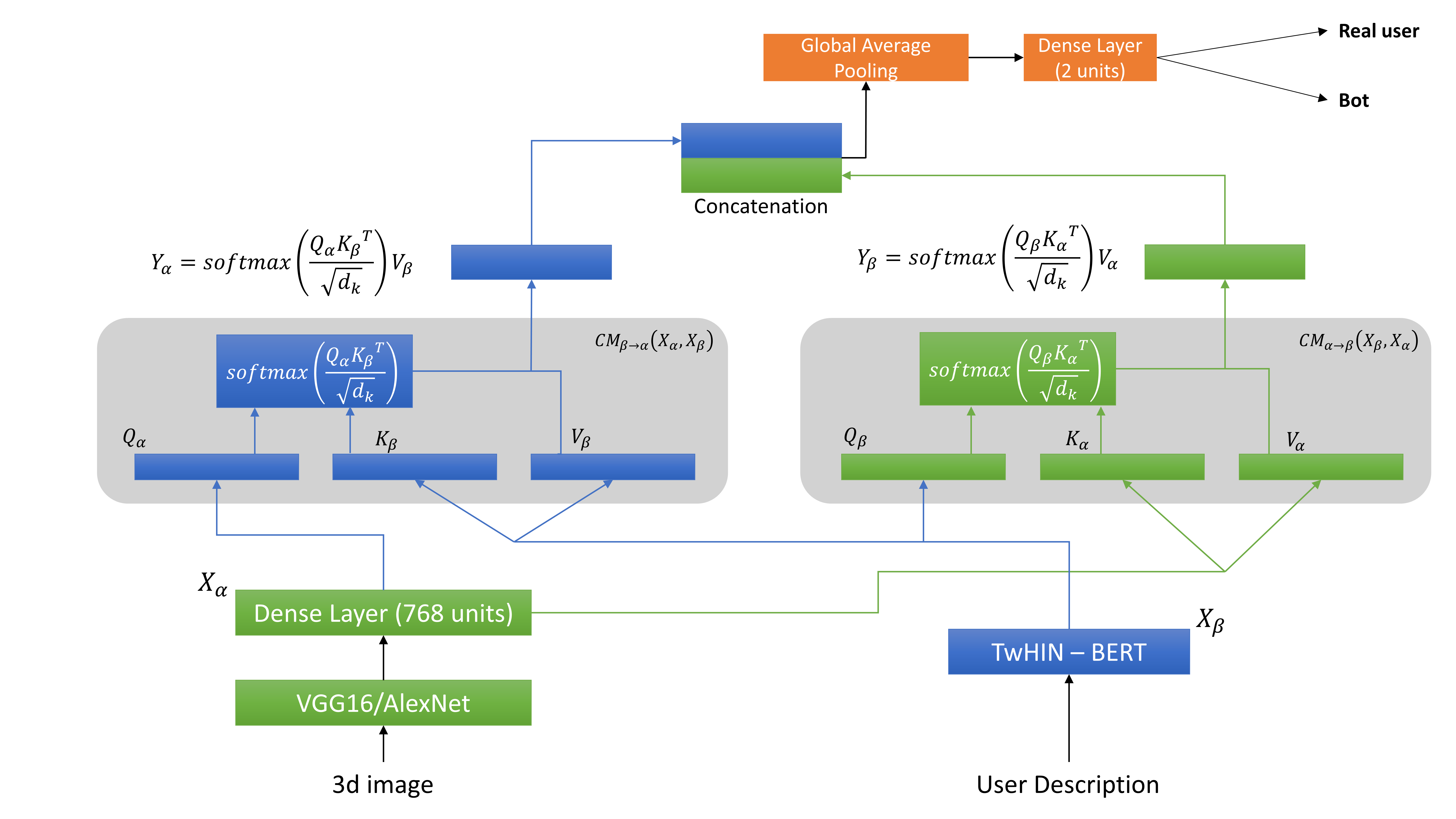}
\caption{Crossmodal Attention}
\label{crossmodal_attention}
\end{subfigure}
\caption{Our introduced approaches}
\label{our_introduced_models_figure}
\end{figure*}

In this section, we describe our proposed models employing both textual and vision modalities. Our introduced models are illustrated in Fig.~\ref{our_introduced_models_figure}.

\subsection{Concatenation}

In terms of the textual modality, we pass the user description field through a TwHIN-BERT model \cite{zhang2022twhin} and extract the [CLS] token. Let $f^t \in \mathbb{R}^{d_t}$ denote the representation vector of the textual modality. $d_t$ denotes the dimensionality and is equal to 768. In terms of the visual modality, we create a 3d image as described in Section~\ref{unimodal_images}. Then, we exploit our best performing model, namely VGG16 for Cresci'17 dataset and AlexNet for TwiBot-20 dataset, as mentioned in Section~\ref{results_unimodal_images}. Specifically, we remove the last layer of VGG16 (and AlexNet) and replace the next-to-last dense layer consisting of 4096 units with one dense layer consisting of 768 units. Let $f^v \in \mathbb{R}^{d_v}$ represent the visual representation vector. $d_v$ is equal to 768. 

After having calculated $f^t$ and $f^v$, we concatenate these two representation vectors into a single vector as following:

\begin{equation}
    z = [f^t, f^v]
\end{equation}
, where $z \in \mathbb{R}^d$. $d=d_t + d_v$ and is equal to 1536.

After this, we pass $z$ through a dense layer consisting of 128 units with a ReLU activation function. Finally, we use a dense layer consisting of two units, which gives the final prediction.

Our introduced model is illustrated in Fig.~\ref{concatenation}.

\subsection{Gated Multimodal Unit}

In terms of the textual modality, we pass the user description field through a TwHIN-BERT model and extract the [CLS] token. Let $f^t \in \mathbb{R}^{d_t}$ denote the representation vector of the textual modality. $d_t$ denotes the dimensionality and is equal to 768. In terms of the visual modality, we create a 3d image as described in Section~\ref{unimodal_images}. Then, we exploit our best performing model, namely VGG16 for Cresci'17 dataset and AlexNet for TwiBot-20 dataset, as mentioned in Section~\ref{results_unimodal_images}. Specifically, we remove the last layer of VGG16 (and AlexNet) and replace the next-to-last dense layer consisting of 4096 units with one dense layer consisting of 768 units. Let $f^v \in \mathbb{R}^{d_v}$ represent the visual representation vector. $d_v$ is equal to 768. 

Next, we use a gated multimodal unit introduced in \cite{arevalo2020gated} for controlling the information flow of the textual and visual modalities. The equations governing the gated multimodal unit are presented below:

\begin{equation}
    h^t = \tanh{(W^t f^t + b^t)}
\end{equation}
\vspace{-2em}
\begin{equation}
    h^v = \tanh{(W^v f^v + b^v)}
\end{equation}
\vspace{-2em}
\begin{equation}
    z = \sigma(W^z [f^t;f^v] + b^z)
\end{equation}
\vspace{-2em}
\begin{equation}
    h = z * h^t + (1-z)*h^v
\end{equation}
\vspace{-2em}
\begin{equation}
    \Theta	= \{W^t, W^v, W^z\}
\end{equation}
, where $h$ is a weighted combination of the textual and visual information $h^t$ and $h^v$ respectively. $\Theta$ denotes the parameters to be learned, while $[;]$ indicates the concatenation operation.

Finally, we pass $h$ through a dense layer consisting of two units for getting the final prediction.

Our introduced model is illustrated in Fig.~\ref{gated_mu}.

\subsection{Crossmodal Attention}

In terms of the textual modality, we pass the user description field through a TwHIN-BERT model. Let $f^t \in \mathbb{R}^{N \times d_t}$ denote the textual representation. $N$ indicates the sequence length, while $d_t$ denotes the dimensionality and is equal to 768. In terms of the visual modality, we create a 3d image as described in Section~\ref{unimodal_images}. Next:

\begin{itemize}
    \item In terms of the Cresci'17 dataset, we exploit our best performing model, namely VGG16, as mentioned in Section~\ref{results_unimodal_images}. Specifically, we get the output of the last CNN layer (after max pooling) as the output of the pretrained VGG16 model. Let $f^v \in \mathbb{R}^{T \times d_v}$ represent the visual representation vector. $T$ and $d_v$ are equal to 64 and 512 respectively. Next, we pass $f^v$ through a dense layer consisting of 768 units.
    \item In terms of the TwiBot-20 dataset, we exploit our best performing model, namely AlexNet, as mentioned in Section~\ref{results_unimodal_images}. Specifically, we get the output of the last CNN layer (after max pooling) as the output of the pretrained AlexNet model. Let $f^v \in \mathbb{R}^{T \times d_v}$ represent the visual representation vector. $T$ and $d_v$ are equal to 49 and 256 respectively. Next, we pass $f^v$ through a dense layer consisting of 768 units.
\end{itemize}

Motivated by \cite{tsai-etal-2019-multimodal,ILIAS2023101485,sanchez-villegas-aletras-2021-point}, we exploit two crossmodal attention layers, i.e., one from textual $f^t$ to visual features $f^v$ and one from visual to textual features.

Specifically, we calculate the scaled dot attention \cite{10.5555/3295222.3295349} as follows:

\begin{equation}
    z = softmax \left(\frac{QK^T}{\sqrt{d}}\right)V
\end{equation}
, where the textual modality corresponds to the query (Q), while the visual modality corresponds to the key (K) and value (V).

Similarly, we design another scaled dot attention layer, where the visual modality corresponds to the query (Q), while the textual modality corresponds to the key (K) and value (V). Let $y$ be the output of the scaled dot attention layer.

Next, we concatenate $z$ and $y$ and pass the resulting matrix into a global average pooling layer. 

Finally, we exploit a dense layer consisting of two units for getting the final output.

Our introduced model is illustrated in Fig.~\ref{crossmodal_attention}.

\section{Experiments}

\subsection{Baselines}

\begin{itemize}
    \item Comparison with state-of-the-art approaches
    \begin{itemize}
        \item  Cresci'17 Dataset
    \begin{itemize}
        \item DeepSBD \cite{9505695}: This method extracts profile, temporal, activity, and content information. Finally, a deep neural network is trained consisting of LSTMs, CNNs, and an attention layer.
        \item DNNBD \cite{KUDUGUNTA2018312}: This method uses only profile information along with SMOTE. We report the results reported in \cite{9505695}. 
        \item DBDM \cite{10.1145/3132847.3133050}: This method models social behavior and content information. A deep neural network consisting of CNNs and BiLSTMs is trained. We report the results reported in \cite{9505695}.
        \item DeBD \cite{8600029}: This method passes the tweet content and the relationship between them into a CNN. Secondly, it uses LSTM to extract the potential temporal features of the tweet metadata. Finally, the authors concatenate the temporal features with the joint content features for detecting social bots. We report the results reported in \cite{9505695}.
        \item MulBot-Glob\_Hier \cite{10020363}: This method extracts multidimensional temporal features from user's timeline, employs dimensionality reduction algorithms, extracts global features, and performs the Agglomerative Hierarchical clustering algorithm.
        \item DNA - sequence (supervised) \cite{7876716}: This method extracts the digital DNA sequence per user by using either the type or content of tweets. Then, the authors leverage the longest common substring (LCS) curves for detecting the social spambots.
        \item Ahmed\_DBSCAN \cite{10020363, AHMED20131120}: This method adopts the approach proposed by \cite{AHMED20131120} and uses DBSCAN to increase the performance. We report the results reported in \cite{10020363}.
        \item Ahmed and Abulaish \cite{AHMED20131120}: This approach exploits the Euclidean distance between feature vectors to create a similarity graph of the accounts. Next, graph clustering and community detection algorithms are used for identifying groups of similar accounts in the graph. We report the results reported in \cite{7876716}.
        \item  Botometer \cite{10.1145/2872518.2889302}: Botometer is a publicly available bot detection tool via the website, Python, or REST APIs. This approach is trained with more than 1,000 features using a Random Forest classifier.
    \end{itemize}
    \item  TwiBot-20 dataset
    \begin{itemize}
        \item Kudugunta and Ferrara \cite{KUDUGUNTA2018312}: This method utilizes user metadata features and tweet content features for identifying bots.
        \item Wei and Nguyen \cite{9014365}: This method uses GloVe embeddings and trains a deep neural network consisting of three layers of BiLSTMs.
        \item Miller et al. \cite{MILLER201464}: Bot detection task is considered as an anomaly detection problem. Specifically, this method extracts 107 features and modifies two stream clustering algorithms, namely StreamKM++ and DenStream.  
        \item Cresci et al. \cite{7436643}: This method extracts the digital DNA sequence per user by using either the type or content of tweets. Then, the authors leverage the longest common substring (LCS) curves for detecting bots in groups. 
        \item Botometer \cite{10.1145/2872518.2889302}.
        \item Alhosseini et al. \cite{10.1145/3308560.3316504}: This method extracts a set of features per user, including age, favourites\_count, statuses\_count, friends\_count, followers\_count, and account length name. Then, a Graph Convolutional Neural Network is trained.
    \end{itemize}
    \end{itemize}
    \item TwHIN-BERT using only the user description field
    \item Our best performing model described in Section~\ref{unimodal_images}
    
\end{itemize}

\subsection{Experimental Setup} 

All the details are provided in Section~\ref{section_experimental_setup}. We use the TwHIN-BERT-base version from the Transformers library in Python \cite{wolf-etal-2020-transformers}.

\subsection{Evaluation Metrics} 

We use the evaluation metrics, which are described in Section~\ref{section_evaluation_metrics}.

\section{Results}

The results of our proposed approaches are reported in Tables \ref{results_multimodal} and \ref{results_multimodal_twibot}. Specifically, Table~\ref{results_multimodal} reports the results on the Cresci'17 dataset, while Table~\ref{results_multimodal_twibot} reports the results on the TwiBot-20 dataset.

\subsection{ Cresci'17 Dataset}
The results of our proposed approaches described in Section~\ref{proposed_multimodal} are reported in Table~\ref{results_multimodal}.

\begin{table}[!htb]
\tiny
\centering
\caption{Performance comparison among proposed models and state-of-the-art approaches on the Cresci'17 dataset. Reported values are mean $\pm$ standard deviation. Results are averaged across five runs. Best results per evaluation metric are in bold.}
\begin{tabular}{lccccc}
\toprule
\multicolumn{1}{l}{}&\multicolumn{5}{c}{\textbf{Evaluation metrics}}\\
\cline{2-6} 
\multicolumn{1}{l}{\textbf{Architecture}}&\textbf{Precision}&\textbf{Recall}&\textbf{F1-score}&\textbf{Accuracy}&\textbf{Specificity}\\
\midrule
\multicolumn{6}{>{\columncolor[gray]{.8}}l}{\textbf{Comparison with state-of-the-art approaches}} \\
DeepSBD \cite{9505695} & \textbf{100.00} & - & 99.81 & 99.83 & - \\
DNNBD \cite{KUDUGUNTA2018312} & 77.66 & - & 75.63 & 78.20 & - \\
DBDM \cite{10.1145/3132847.3133050} & \textbf{100.00} & - & 98.82 & 99.32 & -\\
DeBD \cite{8600029} & 97.73 & - & 97.59 & 97.74 & - \\
MulBot-Glob\_Hier \cite{10020363} & 99.50 & 99.50 & 99.00 & 99.30 & - \\
DNA - sequence (supervised) \cite{7876716} & 98.20 & 97.70 & 97.70 & 97.70 & 98.10 \\
Ahmed\_DBSCAN \cite{10020363, AHMED20131120}& 93.00 & 93.00 & 93.00 & 92.80 & - \\
Ahmed and Abulaish \cite{AHMED20131120} & 94.50 & 94.40 & 94.40 & 94.30 & 94.50 \\
 Botometer \cite{10.1145/2872518.2889302} & - & - & 97.31 & 95.97 & - \\
\midrule
\multicolumn{6}{>{\columncolor[gray]{.8}}l}{\textbf{Unimodal approaches (only user description)}} \\
\textit{TwHIN-BERT} & 99.59 & 99.18 & 99.38 & 99.37 & 99.56 \\
& $\pm$0.50 & $\pm$0.75 & $\pm$0.39 & $\pm$0.40 & $\pm$0.54 \\

\midrule
\multicolumn{6}{>{\columncolor[gray]{.8}}l}{\textbf{Unimodal approaches (only images)}} \\
\textit{VGG16 (type of tweets)} & 99.78 & 99.55 & 99.67 & 99.68 & 99.80 \\
& $\pm$0.44 & $\pm$0.54 & $\pm$0.44 & $\pm$0.42 & $\pm$0.41 \\ \hline
\textit{VGG16 (content of tweets)} & \textbf{100.00} & 99.78 & 99.89 & 99.89 & \textbf{100.00} \\
& $\pm$0.00 & $\pm$0.43 & $\pm$0.22 & $\pm$0.21 & $\pm$0.00 \\ 

\midrule
\multicolumn{6}{>{\columncolor[gray]{.8}}l}{\textbf{Proposed Transformer-based models (images based on the type of tweets)}} \\
\textit{TwHIN-BERT + VGG16} & 99.78 & 99.34 & 99.56 & 99.58 & 99.80 \\
\textit{(Concatenation)} & $\pm$0.44 & $\pm$0.88 & $\pm$0.54 & $\pm$0.52 & $\pm$0.41 \\ \hline
\textit{TwHIN-BERT + VGG16} & 99.79 & 99.79 & 99.79 & 99.79 & 99.78 \\
\textit{(GMU)}& $\pm$0.42 & $\pm$0.42 & $\pm$0.42 & $\pm$0.42 & $\pm$0.43 \\ \hline
\textit{TwHIN-BERT + VGG16} & \textbf{100.00} & 99.79 & 99.90 & 99.89 & \textbf{100.00} \\
\textit{(Cross-Modal Attention)} & $\pm$0.00 & $\pm$0.41 & $\pm$0.21 & $\pm$0.21 & $\pm$0.00 \\ 
\midrule
\multicolumn{6}{>{\columncolor[gray]{.8}}l}{\textbf{Proposed Transformer-based models (images based on the content of the tweets)}} \\
\textit{TwHIN-BERT + VGG16} & 99.77 & 99.77 & 99.77 & 99.79 & 99.80 \\
\textit{(Concatenation)}& $\pm$0.46 & $\pm$0.46 & $\pm$0.46 & $\pm$0.42 & $\pm$0.39 \\ \hline
\textit{TwHIN-BERT + VGG16} & \textbf{100.00} & 99.58 & 99.79 & 99.79 & \textbf{100.00} \\
\textit{(GMU)}& $\pm$0.00 & $\pm$0.52 & $\pm$0.26 & $\pm$0.26 & $\pm$0.00 \\ \hline
\textit{TwHIN-BERT + VGG16} & \textbf{100.00} & \textbf{99.96} & \textbf{99.98} & \textbf{99.98} & \textbf{100.00} \\
\textit{(Cross-Modal Attention)}& $\pm$0.00 & $\pm$0.08 & $\pm$0.04 & $\pm$0.04 & $\pm$0.00 \\ 
\bottomrule
\end{tabular}
\label{results_multimodal}
\end{table}

Regarding our proposed approaches (transformer-based models based on the type of tweets), one can observe that TwHIN-BERT + VGG16 (Cross-Modal Attention) constitutes the best performing model achieving an Accuracy of 99.89\%, a Precision of 100.00\%, a Recall of 99.79\%, a F1-score of 99.90\%, and a Specificity of 100.00\%. It outperforms the other introduced models in Accuracy by 0.10-0.31\%, in Precision by 0.21-0.22\%, in F1-score by 0.11-0.34\%, and in Specificity by 0.20-0.22\%. Also, it outperforms both TwHIN-BERT and VGG16 (type of tweet). Specifically, it surpasses TwHIN-BERT in Precision by 0.41\%, in Recall by 0.61\%, in F1-score by 0.52\%, in Accuracy by 0.52\%, and in Specificity by 0.44\%. It outperforms also VGG16 (type of tweet) in Precision by 0.22\%, in Recall by 0.24\%, in F1-score by 0.23\%, in Accuracy by 0.21\%\%, and in Specificity by 0.20\%. Additionally, TwHIN-BERT + VGG16 (GMU) outperforms TwHIN-BERT + VGG16 (Concatenation) in Precision by 0.01\%, in Recall by 0.45\%, in F1-score by 0.23\%, and in Accuracy by 0.22\%. Also, it outperforms both TwHIN-BERT and VGG16 (type of tweets) in Accuracy by 0.42\% and 0.11\% respectively. In terms of TwHIN-BERT + VGG16 (Concatenation), one can observe that this model obtains worse performance than VGG16 (type of tweets). Specifically, VGG16 (type of tweets) outperforms TwHIN-BERT (Concatenation) in Recall by 0.21\%, in F1-score by 0.11\%, and in Accuracy by 0.10\%. We speculate that this decrease in performance is attributable to the concatenation operation, which assigns equal importance to each modality ignoring the inherent correlations between the two modalities. Overall, we believe that TwHIN-BERT + VGG16 (Cross-Modal Attention) obtains better performance than the other two multimodal models, since it captures the crossmodal interactions. On the other hand, the Gated Multimodal Unit controls the information flow from each modality, while the concatenation operation neglects the inherent correlations.

In terms of our proposed transformer-based models (images based on the content of tweets), one can observe that TwHIN-BERT + VGG16 (Cross-Modal Attention) constitutes the best performing model outperforming both TwHIN-BERT + VGG16 (GMU) and TwHIN-BERT + VGG16 (Concatenation). Specifically, TwHIN-BERT + VGG16 (Cross-Modal Attention) outperforms TwHIN-BERT + VGG16 (GMU) in Recall by 0.42\%, in F1-score by 0.19\%, and in Accuracy by 0.19\%. Although equal Precision scores are achieved, TwHIN-BERT + VGG16 (Cross-Modal Attention) yields a better F1-score, which constitutes the weighted average of Precision and Recall. Additionally, TwHIN-BERT + VGG16 (Cross-Modal Attention) outperforms TwHIN-BERT + VGG16 (Concatenation) in Precision by 0.23\%, in Recall by 0.19\%, in F1-score by 0.21\%, in Accuracy by 0.19\%, and in Specificity by 0.20\%. In comparison with unimodal approaches employing either text or images, we observe that TwHIN-BERT + VGG16 (Crossmodal Attention) outperforms both TwHIN-BERT and VGG16 (content of tweets). Finally, TwHIN-BERT + VGG16 (Cross-Modal Attention) with images based on the content of tweets outperforms all the introduced models exploiting images based on the type of tweets. Therefore, TwHIN-BERT + VGG16 (Cross-Modal Attention) with images based on the content of tweets constitutes our best performing model.

In comparison with the state-of-the-art approaches, one can observe that our best performing model, namely TwHIN-BERT + VGG16 (Cross-Modal Attention) with images based on the content of the tweets, outperforms these approaches in Precision by 0.50-22.34\% (except for DeepSBD and DBDM), in Recall by 0.46-6.96\%, in F1-score by 0.17-24.35\%, in Accuracy by 0.15-21.78\%, and in Specificity by 1.90-5.50\%. Similarly, TwHIN-BERT + VGG16 (Cross-Modal Attention) with images based on the type of tweets, surpasses the existing research initiatives in Precision by 0.50-22.34\% (except for DeepSBD and DBDM), in Recall by 0.29-6.79\%, in F1-score by 0.09-24.27\%, in Accuracy by 0.06-21.69\%, and in Specificity by 1.90-5.50\%. Although our best performing model outperforms DeepSBD by a small margin of 0.17\% and 0.15\% in F1-score and Accuracy respectively, our best performing model has multiple advantages over DeepSBD. Firstly, DeepSBD processes the tweets of each user, extracts GloVe embeddings, and creates a 3d matrix which is given as input to a 6-layer CNN increasing the computational demands. Additionally, this method extracts a set of features per user. On the contrary, our method seems to be simpler and more effective, since it does not rely on a feature extraction strategy; it relies only on the user description and the sequence of actions performed by the user. Additionally, the authors use GloVe embeddings instead of using a language model based on transformers. Thus, this approach inherits the limitations of the GloVe embeddings. In terms of the training time per epoch, 120 to 180 seconds are required for training by our model, while the training time of DeepSBD ranges from 248 to 634 seconds.

\subsection{TwiBot-20 Dataset}

The results of our proposed approaches described in Section~\ref{proposed_multimodal} are reported in Table~\ref{results_multimodal_twibot}.

In terms of the proposed transformer-based models using images based on the type of tweets posted by the users, one can observe that TwHIN - BERT + AlexNet (Cross-Modal Attention) constitutes the best performing model in terms of Recall, F1-score, and Accuracy. Specifically, it outperforms the other introduced multimodal models in Recall by 2.55-2.85\%, in F1-score by 0.83-1.26\%, and in Accuracy by 0.28-0.76\%. Additionally, TwHIN-BERT + AlexNet (Cross-Modal Attention) with images based on the type of tweets, surpasses the performance achieved by unimodal models, i.e., TwHIN-BERT (using user description) and AlexNet (using images). Similar differences in models' performance are observed, when the images based on the content of tweets are used as input to proposed approaches. To be more precise, the usage of the crossmodal attention as a fusion method yields the best results surpassing the performance achieved by concatenation and gated multimodal unit as fusion approaches. TwHIN-BERT + AlexNet (Cross-modal Attention) outperforms the other multimodal models in Recall, F1-score, and Accuracy by 3.16-4.43\%, 1.26-2.03\%, and 0.70-1.36\% respectively. TwHIN-BERT + AlexNet (Cross-Modal Attention) outperforms also the unimodal models employing either text or images. Specifically, it surpasses TwHIN-BERT in F1-score and Accuracy by 3.18\% and 3.25\% respectively, while it also outperforms AlexNet in F1-score and Accuracy by 9.50\% and 7.47\% respectively. Overall, we observe that the usage of multiple modalities improves bots' detection performance. Cross-modal Attention boosts performance, since it models the cross-modal interactions. On the other hand, the gated multimodal unit refers to a weighting strategy, which controls the contribution of each modality to the bots' prediction task. The concatenation operation is not capable of capturing the interactions of the two modalities or controlling the contribution of each modality, since equal importance to text and image information is assigned. 

In comparison with state-of-the-art approaches, our best performing model outperforms these approaches in Accuracy and F1-score by 3.40-26.73\% and 0.97-65.58\% respectively. We observe that although Botometer was capable of identifying the social spambots in Cresci'17 dataset, the performance drops significantly in TwiBot-20 dataset, which verifies the fact that bots in Twitter have changed their behavior and have managed to evade previous detection methods. Similarly, we observe that the other approaches employing feature extraction techniques fail to detect bots. On the contrary, our approach relying only on the user description and the sequence of actions performed by the user based on the content of tweets seems to be an effective method.

\begin{table}[!htb]
\tiny
\centering
\caption{Performance comparison among proposed models and state-of-the-art approaches on the TwiBot-20 dataset. Reported values are mean $\pm$ standard deviation. Results are averaged across five runs. Best results per evaluation metric are in bold.}

\begin{tabular}{lccccc}
\toprule
\multicolumn{1}{l}{}&\multicolumn{5}{c}{\textbf{Evaluation metrics}}\\
\cline{2-6} 
\multicolumn{1}{l}{\textbf{Architecture}}&\textbf{Precision}&\textbf{Recall}&\textbf{F1-score}&\textbf{Accuracy}&\textbf{Specificity}\\
\midrule
\multicolumn{6}{>{\columncolor[gray]{.8}}l}{\textbf{Comparison with state-of-the-art approaches}} \\
Kudugunta and Ferrara \cite{KUDUGUNTA2018312} &  - & - & 47.26 & 59.59 & - \\
Wei and Nguyen \cite{9014365} & - & - & 75.33 & 71.26 & - \\
Miller et al. \cite{MILLER201464} & - & - & 62.66 & 48.01 & - \\
Cresci et al. \cite{7436643} & - & - & 10.72 & 47.93 & -\\
Botometer \cite{10.1145/2872518.2889302} & - & - & 48.92 & 55.84 & - \\
Alhosseini et al. \cite{10.1145/3308560.3316504} & - & - & 73.18 & 68.13 & - \\
\midrule
\multicolumn{6}{>{\columncolor[gray]{.8}}l}{\textbf{Unimodal approaches (only user description)}} \\
\textit{TwHIN-BERT} & 71.41 & 75.12 & 73.12 & 71.41 & 67.41 \\
& $\pm$1.64 & $\pm$4.15 & $\pm$1.61 & $\pm$1.07 & $\pm$3.76 \\

\midrule
\multicolumn{6}{>{\columncolor[gray]{.8}}l}{\textbf{Unimodal approaches (only images)}} \\
\textit{AlexNet (type of tweets)} & 68.33 & 65.89 & 67.05 & 66.41 & 66.96 \\
& $\pm$1.26 & $\pm$2.05 & $\pm$0.72 & $\pm$0.62 & $\pm$2.79 \\ \hline
\textit{AlexNet (content of tweets)} & 70.36 & 63.64 & 66.80 & 67.19 & 71.01 \\
& $\pm$1.11 & $\pm$1.53 & $\pm$0.37 & $\pm$0.28 & $\pm$2.20 \\ 

\midrule
\multicolumn{6}{>{\columncolor[gray]{.8}}l}{\textbf{Proposed Transformer-based models (images based on the type of tweets)}} \\
\textit{TwHIN-BERT + AlexNet} & 74.22 & 75.01 & 74.52 & 73.42 & 71.70 \\
\textit{(Concatenation)}& $\pm$1.93 & $\pm$3.35 & $\pm$0.88 & $\pm$0.53 & $\pm$3.98 \\ \hline
\textit{TwHIN-BERT + AlexNet} & 74.70 & 75.31 & 74.95 & 73.90 & \textbf{72.39} \\
\textit{(GMU)}& $\pm$1.56 & $\pm$3.02 & $\pm$1.26 & $\pm$1.00 & $\pm$2.91 \\ \hline
\textit{TwHIN-BERT + AlexNet} & 73.98 & 77.86 & 75.78 & 74.18 & 70.20 \\
\textit{(Cross-Modal Attention)}& $\pm$2.40 & $\pm$3.02 & $\pm$0.48 & $\pm$0.78 & $\pm$4.59 \\ 
\midrule
\multicolumn{6}{>{\columncolor[gray]{.8}}l}{\textbf{Proposed Transformer-based models (images based on the content of the tweets)}} \\
\textit{TwHIN-BERT + AlexNet} & 74.31 & 74.26 & 74.27 & 73.30 & 72.27 \\
\textit{(Concatenation)}& $\pm$0.90 & $\pm$1.39 & $\pm$0.35 & $\pm$0.29 & $\pm$1.78 \\ \hline
\textit{TwHIN-BERT + AlexNet} & \textbf{74.74} & 75.53 & 75.04 & 73.96 & 72.27 \\
\textit{(GMU)}& $\pm$1.88 & $\pm$3.58 & $\pm$0.93 & $\pm$0.31 & $\pm$3.93 \\ \hline
\textit{TwHIN-BERT + AlexNet} & 74.16 & \textbf{78.69} & \textbf{76.30} & \textbf{74.66} & 71.61 \\
\textit{(Cross-Modal Attention)}& $\pm$1.14 & $\pm$2.97 & $\pm$0.83 & $\pm$0.31 & $\pm$3.16 \\ 
\bottomrule
\end{tabular}
\label{results_multimodal_twibot}
\end{table}

 \section{Qualitative and Error Analysis}

In this section, we perform a qualitative and error analysis of the best performing model on the TwiBot-20 dataset, namely TwHIN-BERT + AlexNet (Cross-Modal Attention) using images based on the content of tweets. Table~\ref{error_analysis} reports the predictions on some sample instances. Specifically, rows 1 and 2 refer to correct classifications as real users. Similarly, rows 3-7 refer to correct predictions as bots. Rows 8-10 refer to incorrect predictions made by our best performing model. Specifically. our model predicts these instances as belonging to a bot, while the real label corresponds to a genuine account. Similarly, rows 11 and 12 indicate incorrect classifications, where the actual label is a bot, while the predicted label is a real user.

Regarding correct classifications corresponding to genuine accounts, we observe that the Digital DNA of a real user includes a great number of consecutive mentions. Thus, our model learns that consecutive mentions indicate a real user, since mentions indicate often interactions with other users. Additionally, we observe that the user description field refers to a simple description of the biography of the user. 

In terms of the correct classifications referring to bots, we observe in row 3 that the digital DNA includes a lot of consecutive hashtags. Although Digital DNAs of rows 4-7 include a lot of mentions, we observe that the user description fields have common characteristics. Specifically, we observe that the users' descriptions of rows 4-6 include many words separated by commas (,). The user description of row 7 includes many hashtags (8 in number) and emojis. Therefore, both user description and the activity of users are important towards the final prediction.

Next, we examine reasons of misclassifications, i.e., rows 8-10. Specifically, we hypothesize that our model misclassifies row 8 as bot, since the Digital DNA consists of a great number of consecutive URLs. Also, we believe that row 9 is predicted as bot, since the user description includes many mentions. Finally, we observe that row 10 is similar to rows 4-6, and thus this row is predicted falsely as bot. Specifically, we observe that the user description of row 10 includes many words separated by commas.

Finally, we investigate the reasons of misclassifications in rows 11 and 12. We observe that Digital DNAs contain a lot of consecutive mentions, which refer to a genuine account (see rows 1 and 2). This verifies the fact that bots have found ways to evade detection approaches and mimic human behavior.  

\begin{table*}
\caption{ Sample Instances for Analysis}

\begin{tabularx}{\textwidth}{>{\hsize=.01\hsize}X|>{\hsize=.40\hsize}X|X|>{\hsize=.03\hsize}X|>{\hsize=.01\hsize}X}
\toprule
\multicolumn{1}{c|}{\textbf{ID}} & \multicolumn{1}{c|}{\textbf{User Description}} & \multicolumn{1}{c|}{\textbf{Digital DNA}} & \multicolumn{1}{c|}{\textbf{Real}} & \multicolumn{1}{c}{\textbf{Pred.}} \\ \midrule
1 & Sports Broadcaster \& Event Host  @skysports & 
MXMMMNUUNUMXUMUMMMUUMXXXXUMMMMMMMMMXXUXMMMMXU
NUXXMMUMUMXUUXXXMMMUUXXUXXMUXUMMMUXMXXXUXUMMMUMX
UUXUMMUXXXMMMXMUUUMMXUUMMMMMXXUUMMXUMMMUMMUXXMMMMUMUXXUUMMUMMUU
UMMMUXXXUUNUUMXMMMXMXXUMXMUUUUXMMMMMUUMXMUMX & Real User & Real User \\ \hline
2 & Former NASA and Apple engineer. Current YouTuber and friend of science. \textit{<URL>} & XMUMMMMXMXMMXXMXXMMXMXUMMXMMUXUUMMMMMMXMMMMMM
MMMMXMMMMMUUUMMXMMMMXMMMXXUNXMMUMMMMMMMMUMUXMMNXXUUMXUMXMUMM
MMMMMXMMUUUMMUMMMMMMMUUMMMUMMMMMMMMMNMNMUUXMMXMNMMXMMM
MMXMMXUMUMMMMMMUUNXUMMMMMUMMMMUMMUUXXMXX & Real User & Real User \\ \hline
3 & Fun, outgoing guy that loves to meet new people. Casual gamer with a competitive nature. & HXXXXXMMXXMXMXMMXXHXXHHHHMXXXXXXXXXXXXXMXXXXXXXXU
MHHXXUXMUXXXUXXXXXUHMHMXMMMMXXXHHHMHHHHHXHXHHHHXHHHHHHHHH
HXHHHHHHMHXMMMMXHXHHHMHHHMXHNXHHHMMXXXXHMMMHHHHHHHHMH
HHMXHMHHHNHHHHHMHHHHHHHHHHHHHUHHHXHHHXXHM & Bot & Bot \\ \hline
4 & progressive, vegan, animal rights, social justice,  anti racism, resistance, anti trump, lock him up, the SDNY will get the whole crime family \textit{<raised fist emoji>} \#FBR  & MXMMMMXMMUXMMMMXUMMMMMMMMUMUMMMMUXMXM XMMMMMXMMMMMXMXMMMMMMMMMMMMMMMMUMMMMMM MUXMMXXUMUMMMMMXMMMMMMMMXXMMMMMXXUMMMM MMXMMXUMUMMMMMXMMMMMMMMMXMXMXUMMMMMUXM MMXMMMXMMMXMXMMUXMXMMMUMMMXMMMUMMXUXXM XMMMXMXMMMM & Bot & Bot \\ \hline
5 & wife, mom, gma, Dem, liberal, truth, justice, equality, free press, free speech, LGBTQ, unions, proChoice, singlePayer, science\#Resistance, \#VoteBluetoSaveAmerica  & MXMMMMXMMMMMMMUMMMMMMXMMMMMMXXMMHXMXXXMMMX
MMMMMXXMXMMMMMXMMMXMMXMXXXXMMMMMMMMMMXMMMMMMXMMXMXMMUUMMMMUU
XXMXMMXMXMMMMXMMMMMMMMMUMXMMXMMMMUUMUMMXMMUXMXMUM MMMMXMMMMUMXMMMMMXMMMMMMMMXMMMXMMMMMMMMMMXMXMMXXX & Bot & Bot \\ \hline
6 & Christian, Conservative, Army,Life member NRA,Believe in God my Savior and Lord.Believe in America  and believe in our great State of TEXAS !!!Senator Cruz III\%  & XMMMMXMXMXMXMMMMXMMNMMXMMMXMMXMMXXMXXM MXMMMMMMMMMMMMMMMMXMMMMMMMMMMMXMNMMMUMM MXMMMMMMMXMMMMMXMMMMMMUXMMMUMXMMMUMMMM XMXMMMMMMMXMXMMMMMMMMMMMMMMMXMMMMMMMMMM MXMXMMXMMMMMXMMMMMMMMXMMM & Bot & Bot \\ \hline
7 & I AM HERE 4 POTUS \#GGOAT \#MomentsMatter \textit{<glowing star emoji>} \#BELIEVING=PEACE  \textit{<folded hands emoji>} \textit{<paw prints emoji>} \#BeachBabeBum \textit{<desert island emoji>} \textit{<eyes emoji>} \textit{<fox emoji>} \#TrumpTrain \textit{<locomotive emoji>} \#MAGA \#KAG \#RedWomenRealWomen ArkTrumper \textit{<anchor emoji>} SADLY\_Disabled & MMXXXMMMXXMMMMMMMXXMXMMMMMMMMMMMMMMMMMMXMXM
MMMMNUXMMXMMMXMXMMMMMMMMXMXXXXXXMXNNNMUMXMMXXXXXXMMM
XMMMMMXMMXXMMMMMMXXMMMMXXMMMMMMXMMXMMMXXMXXMMXXMMXMMMM
MMMMMXXMMMMMMMMMMMMMMXXMMMXMMMMXMMMXMXXMMMXMMMMNMMX & Bot & Bot \\ \hline

8 & Pasta lover. I don't tweet much. My new Netflix series Master of None is now streaming on Netflix. I wrote a book called Modern Romance. & XXXXXXXXXUXXXUUUUUUUUUUUUUXUUUUUUUUUUUUUUUUUUUUUXUU
UUUUUUUUUUUUUUUUNUXUXUUUUUUUUXXNNUXXHUUXXXNXXXXXUUUUMNXUUX
MUUUUUUXUNXXUUXUUXXUUMXXUUNXXXUUUUXMXNNUUUUUMUUUUHUUUUUXUUUUU UUUUUUUUUUUUUUUUUUUUUMUUUUUUUU & Real User & Bot \\ \hline
9 & @mu\_foundation @veteransgarage \& @SSAFA ambassador. @helpforheroes @rmchcharity and @manchesterpride patron. & XMUUXXMXUMXUUUUUMUMMNMUMUUMUMUU MUUUMMMNXUXMUMMNUM UMUXXUXXUUMMUUXXNUMMUXUXMXMXXMMXXUNUMMUUUUUMUUUUUM MXMNUMXUUUMXUUUUXXMMXMXMMUUNMMMMXMUUMMMMXNXMMMMMM MXUNMNNNMXXXUXNUNXMUMXXUUMMHUUUUNNXXMMMMMMXUMXUUNMNM & Real User & Bot \\ \hline
10 & news/media, music/guitar, ECB, EU, Frankfurt, Greece, politics. Opinions my own, not necessarily my employer's (=ECB). RT's often mean I agree, but not always. & MXMXXXMXUMMMMUMMUMXMXXXUXXMNUMUUXXMUMXMMMMUXXX
MMXMUXMNMMMUXMMXMMXMMMMMMMMMMXMMUMMMXUNNNNNMMXXXUMMMM
XXMMUXMMXMXMMMXMMMMXMUMMMXMMXMMMXXXMMXMMMMXUXMXX
XXMMXXMMXMXUXXMMMXXXMMMMMMMMMXNXMXXXXMMMXMMMMXM UXXXUU & Real User & Bot \\ \hline
11 & Political Booking Producer at @nbcnews | @todayshow @nbcnightlynews & MMMXNXNXMMMMXMMXMMMMMXMMMMXMMMMMMMMXMXXXXXM
XMMMXMMXMMMXMMMMMMXMXXMMXMMMMMMXMMXXMMMMMXXMMXUMXXMMMMMMMXXMMM MMXMXHMMMMMMMMMMMMXMMMMXMMXXXXNXMMXXMMMXMMMXMMMMMMMMXMMMXNXMXXXMMXMXMM MXMMMXMXMMMXMMMMMMMXNMXMX & Bot & Real User \\ \hline
12 & Lets Hope and pray for The Best & MMMMMMXMMMMMMMMMMMMMMMMMMMMMMMMXMMMXMMUMM
MM & Bot & Real User \\ 
\bottomrule
\end{tabularx}
\label{error_analysis}
\end{table*}

 \section{Discussion}

\subsection{Limitations}

This study has some limitations. Firstly, we did not apply explainability approaches to explain the predictions made by our proposed approaches. Additionally, this study requires labelled data. On the contrary, self-supervised learning approaches address the issue of labels' scarcity. 

\subsection{Scalability and Real-time detection}

An often neglected concern among researchers introducing methods for identifying bots in Twitter is the reliance on the platform APIs \cite{yang2023social}. Specifically, data must be retrieved from social media platforms, so as to ensure that the researchers perform the necessary experiments. 

However, in 2023, Twitter decided to end free access to their APIs. As a result of this modification, several research projects\footnote{https://www.reuters.com/technology/elon-musks-x-restructuring-curtails-disinformation-research-spurs-legal-fears-2023-11-06/} have been cancelled or suspended for a period of time. After that, legal obligations in the bloc’s Digital Services Act (DSA)\footnote{https://techcrunch.com/2023/11/17/change-in-xs-terms-indicate-eu-researchers-will-get-api-access/} require larger platforms, including Meta, Twitter, and Google, to provide data access to external researchers doing public interest research on systemic risks. However, the access to Twitter APIs still remains uncertain. 

Modifications also to the API affect the bot detection algorithms. For example, the field \texttt{geo\_enabled} was removed in 2019 due to privacy reasons. Therefore, algorithms trained with this features, should be modified. 

Data access is inextricably linked with scalability. Scalability refers to the analysis of streaming data with limited computing resources \cite{Yang_Varol_Hui_Menczer_2020}. Specifically, the speed at which a method processes a group of accounts depends highly on the possibility for access to the Twitter API. For instance, although the construction of large graphs leads to detection of bots with increasing evaluation performance, fetching such information is not feasible. The study in \cite{Ferrara_2023} claims that the following methods are required for improving the scalability of bot detection algorithms:

\begin{itemize}
    \item Model compression and distillation
    \item Incremental Learning and online algorithms
    \item Parallel and distributed processing
    \item Stream-based processing and data reduction
\end{itemize}

\subsection{Generalization of our approach on other social media}

The introduced approaches in this paper can be generalized on other social media, including Facebook and Reddit. Similar to Twitter, users on the Facebook platform make posts, reposts, and reply to other users. At the same time, their posts may include mentions to other users, hashtags, and URLs. Additionally, each user may include a description field in his/her profile. 

However, platforms' policy about data access is critical. For instance, investigating bot detection approaches on the Facebook platform is a very difficult task due to the unwillingness of Facebook to share individual account data \cite{yang2023social}.

\section{Conclusion and Future Work}

In this paper, we present the first study introducing multimodal and cross-modal models for detecting bots in Twitter by exploiting only the user description and 3d images, which represent the actions of each user. Firstly, we create two digital DNA sequences based on both the type and content of the tweets each user posts. Next, we apply a DNA-to-image conversion algorithm and create two 3d images per user based on the two digital DNA sequences. Finally, we fine-tune several pretrained models of the vision domain and show that VGG16 achieves the highest evaluation results. Next, we introduce multimodal and crossmodal models. First, we pass each user description field into a TwHIN-BERT and obtain a textual representation. We pass each 3d image through a VGG16 model and obtain a visual representation. Finally, we compare three methods for fusing the textual and visual representations, including the concatenation operation, the gated multimodal unit, and the cross-modal attention. Findings show that the crossmodal attention outperforms the other introduced approaches, be it either unimodal or multimodal approaches, obtaining also comparable performance with the state-of-the-art approaches.

In the future, we aim to apply our models in a federated learning framework. Applying also continual learning approaches is one of our future plans. Finally, we plan to exploit explainability techniques for rendering the introduced approaches explainable.

\bibliographystyle{IEEEtran}
\bibliography{references}


\section{Biography Section}

\vspace{11pt}
\begin{IEEEbiography}[{\includegraphics[width=1in,height=1.25in,clip,keepaspectratio]{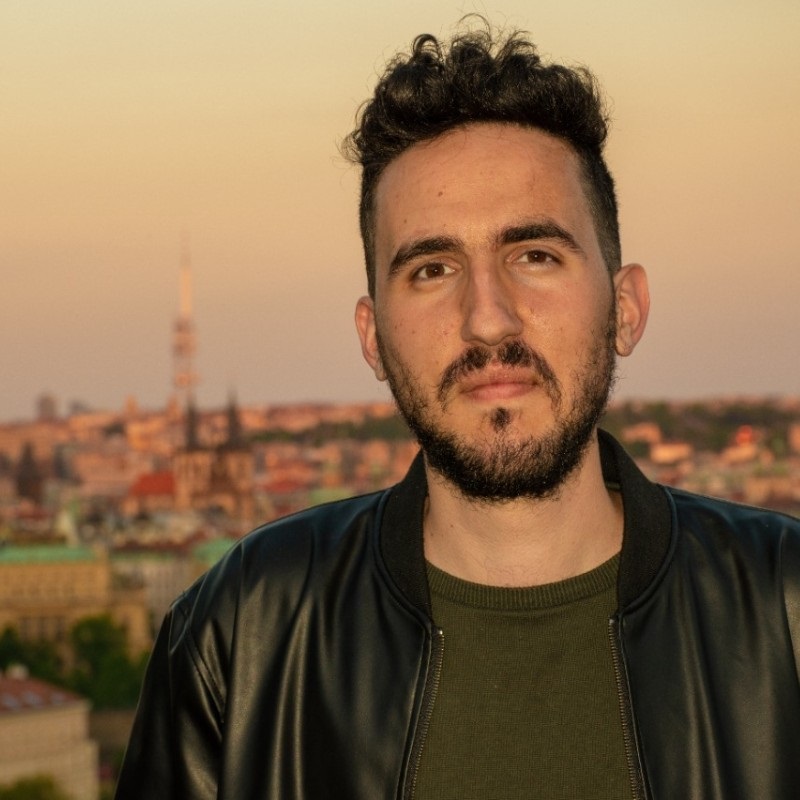}}]{Loukas Ilias}
received the integrated master’s degree
from the School of Electrical and Computer Engineering (SECE), National Technical University of
Athens (NTUA), Athens, Greece, in June 2020,
where he is currently pursuing the Ph.D. degree with
the Decision Support Systems (DSS) Laboratory,
SECE. He has completed a Research Internship
with University College London (UCL), London,
U.K.

He is a Researcher with the DSS Laboratory,
NTUA, where he is involved in EU-funded research
projects. He has published in numerous journals, including IEEE Journal of Biomedical and Health Informatics, IEEE Transactions on Computational Social Systems, Knowledge-Based Systems (Elsevier), Expert Systems With Applications (Elsevier), Applied Soft Computing (Elsevier), Online Social Networks and Media (Elsevier), Computer Speech and
Language (Elsevier), IEEE Access, Frontiers in Aging Neuroscience, and Frontiers in Big Data. His
research has also been accepted for presentation at international conferences,
including the IEEE-EMBS International Conference on Biomedical and Health Informatics (BHI’22)
and the IEEE International Conference on Acoustics, Speech, and Signal
Processing (ICASSP) 2023. His research interests include speech processing,
natural language processing, social media analysis, and the detection of
complex brain disorders.

\end{IEEEbiography}

\vspace{11pt}

\begin{IEEEbiography}[{\includegraphics[width=1in,height=1.25in,clip,keepaspectratio]{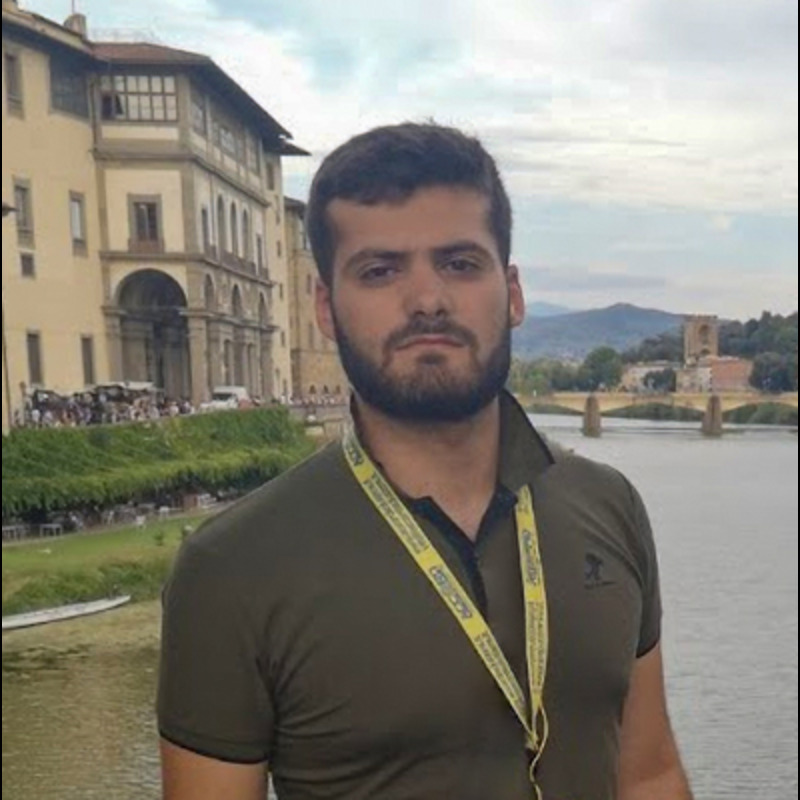}}]{Ioannis Michail Kazelidis}
received the integrated master’s degree
from the School of Electrical and Computer Engineering (SECE), National Technical University of
Athens (NTUA), Athens, Greece, in October 2023. His research interests include deep learning and social media analysis.

\end{IEEEbiography}

\vspace{11pt}

\begin{IEEEbiography}[{\includegraphics[width=1in,height=1.25in,clip,keepaspectratio]{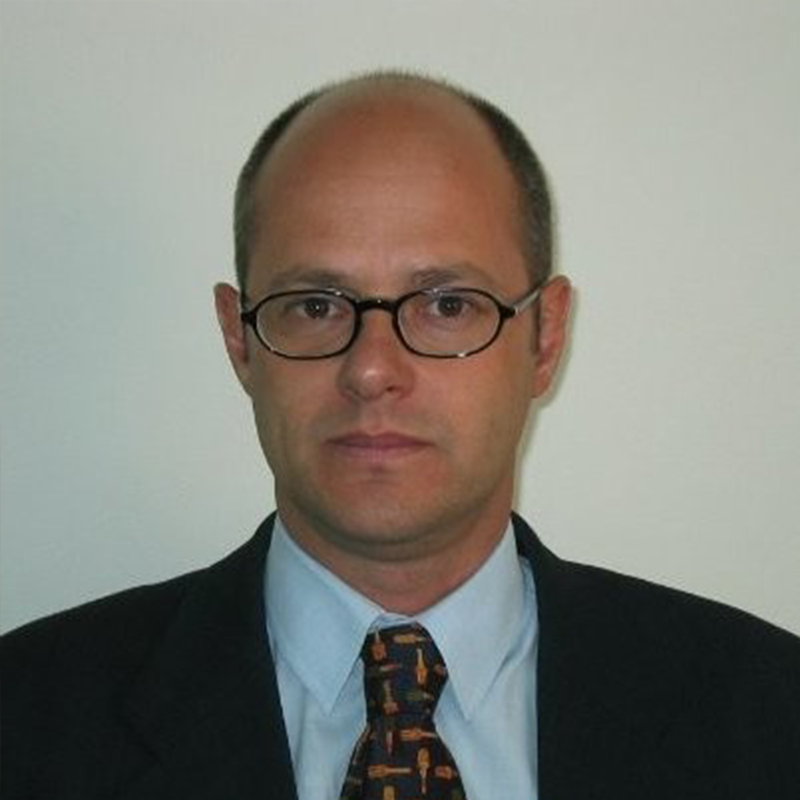}}]{Dimitris Askounis}
was the Scientific Director of
over 50 European research projects in the above
areas (FP7, Horizon2020, and so on). For a number
of years, he was an Advisor to the Minister of
Justice and the Special Secretary for Digital Convergence for the introduction of information and
communication technologies in public administration. Since June 2019, he has been the President
of the Information Society SA, Kallithea, Greece.
He is currently a Professor at the School of Electrical and Computer Engineering, National Technical
University of Athens (NTUA), Athens, Greece, and the Deputy Director
of the Decision Support Systems Laboratory. He has over 25 years of
experience in decision support systems, intelligent information systems and
manufacturing, e-business, e-government, open and linked data, big data
analytics, Artificial Intelligence (AI) algorithms, and the application of modern
Information Technology (IT) techniques in the management of companies and
organizations.

\end{IEEEbiography}

\vfill

\end{document}